\documentclass[10pt,logo,copyright]{nvidiatechreport}
\usepackage[authoryear,round]{natbib}

\usepackage[utf8]{inputenc} %
\usepackage[T1]{fontenc}    %

\usepackage{parskip}        %
\usepackage{url}            %
\usepackage{booktabs}       %
\usepackage{amsfonts}       %
\usepackage{nicefrac}       %
\usepackage{microtype}      %
\usepackage{xcolor}         %
\usepackage[dvipsnames]{xcolor} %
\usepackage{graphicx}
\usepackage{animate}        %
\usepackage{subcaption}
\usepackage{tabularx}
\usepackage{makecell}
\usepackage{adjustbox}
\usepackage{setspace}
\newcolumntype{M}[1]{>{\centering\arraybackslash}m{#1}}
\usepackage{float}
\usepackage{tikz}
\usetikzlibrary{positioning,shapes,arrows}
\usepackage{amsmath,amsfonts,bm, bbm,leftindex}
\usepackage{multirow}
\usepackage{comment}
\usepackage{gensymb}
\usepackage{lipsum}
\usetikzlibrary{arrows.meta, positioning, fit}
\usepackage[para]{threeparttable}
\usepackage{tikz}
\usetikzlibrary{tikzmark}
\usepackage{hyperref}
\usepackage[most]{tcolorbox}
\usepackage{fancyvrb}
\usepackage{fvextra}
\usepackage{dashrule}
\usepackage{wrapfig}

\def\eqref#1{equation~\ref{#1}}

\def\1{\bm{1}}

\DeclareMathAlphabet{\mathsfit}{\encodingdefault}{\sfdefault}{m}{sl}
\SetMathAlphabet{\mathsfit}{bold}{\encodingdefault}{\sfdefault}{bx}{n}

\makeatletter
\let\save@mathaccent\mathaccent
\newcommand*\if@single[3]{%
  \setbox0\hbox{${\mathaccent"0362{#1}}^H$}%
  \setbox2\hbox{${\mathaccent"0362{\kern0pt#1}}^H$}%
  \ifdim\ht0=\ht2 #3\else #2\fi
  }
\newcommand*\rel@kern[1]{\kern#1\dimexpr\macc@kerna}
\newcommand*\widebar[1]{\@ifnextchar^{{\wide@bar{#1}{0}}}{\wide@bar{#1}{1}}}
\newcommand*\wide@bar[2]{\if@single{#1}{\wide@bar@{#1}{#2}{1}}{\wide@bar@{#1}{#2}{2}}}
\newcommand*\wide@bar@[3]{%
  \begingroup
  \def\mathaccent##1##2{%
    \let\mathaccent\save@mathaccent
    \if#32 \let\macc@nucleus\first@char \fi
    \setbox\z@\hbox{$\macc@style{\macc@nucleus}_{}$}%
    \setbox\tw@\hbox{$\macc@style{\macc@nucleus}{}_{}$}%
    \dimen@\wd\tw@
    \advance\dimen@-\wd\z@
    \divide\dimen@ 3
    \@tempdima\wd\tw@
    \advance\@tempdima-\scriptspace
    \divide\@tempdima 10
    \advance\dimen@-\@tempdima
    \ifdim\dimen@>\z@ \dimen@0pt\fi
    \rel@kern{0.6}\kern-\dimen@
    \if#31
      \overline{\rel@kern{-0.6}\kern\dimen@\macc@nucleus\rel@kern{0.4}\kern\dimen@}%
      \advance\dimen@0.4\dimexpr\macc@kerna
      \let\final@kern#2%
      \ifdim\dimen@<\z@ \let\final@kern1\fi
      \if\final@kern1 \kern-\dimen@\fi
    \else
      \overline{\rel@kern{-0.6}\kern\dimen@#1}%
    \fi
  }%
  \macc@depth\@ne
  \let\math@bgroup\@empty \let\math@egroup\macc@set@skewchar
  \mathsurround\z@ \frozen@everymath{\mathgroup\macc@group\relax}%
  \macc@set@skewchar\relax
  \let\mathaccentV\macc@nested@a
  \if#31
    \macc@nested@a\relax111{#1}%
  \else
    \def\gobble@till@marker##1\endmarker{}%
    \futurelet\first@char\gobble@till@marker#1\endmarker
    \ifcat\noexpand\first@char A\else
      \def\first@char{}%
    \fi
    \macc@nested@a\relax111{\first@char}%
  \fi
  \endgroup
}
\makeatother

\newcommand{\traj}{\bm{\tau}}
\newcommand{\dyntraj}{\bm{a}}
\newcommand{\observation}{\bm{o}}
\newcommand{\minade}{\text{minADE}_{6}}
\newcommand{\reasoning}{\textsc{Reason}}

\usepackage{pifont}

\usepackage[nameinlink]{cleveref}
\crefname{equation}{Eq.}{Eqs.}
\crefname{figure}{Fig.}{Figs.}
\crefname{section}{Sec.}{Sec.}
\crefname{appendix}{App.}{App.}
\crefname{table}{Tab.}{Tabs.}
\crefname{algorithm}{Algo}{Algo}
\crefname{thm}{Thm}{Thm}
\Crefname{thm}{Thm}{Thm}
\crefname{prop}{Prop}{Prop}
\makeatletter
\AddToHook{cmd/appendix/before}{\def\cref@section@alias{appendix}}
\makeatother

\usepackage{makecell}
\usepackage[table]{xcolor}

\newif\ifshowtodos
\showtodostrue     %
\ifshowtodos
  \newcommand{\todo}[1]{\textcolor{red}{[\textit{TODO: #1}]}\xspace}
\else
  \newcommand{\todo}[1]{}
\fi

\definecolor{darkred}{rgb}{0.7, 0.0, 0.0}

\newcommand{\reasoningvla}{Alpamayo-R1\xspace}
\newcommand{\reasoningvlashort}{AR1\xspace}

\newcommand{\datafullnamenosp}{Chain of Causation}
\newcommand{\datafullname}{\datafullnamenosp~}
\newcommand{\datafullnamelowernosp}{chain-of-causation}
\newcommand{\datafullnamelower}{\datafullnamelowernosp~}
\newcommand{\datashortnamenosp}{CoC}
\newcommand{\datashortname}{\datashortnamenosp~}

\newcommand{\crefnames}[3]{%
  \@for\next:=#1\do{%
    \expandafter\crefname\expandafter{\next}{#2}{#3}%
  }%
}

\newcommand\mypara[1]{\vspace{1.5mm}\noindent\textbf{#1}}

\usepackage[most]{tcolorbox}
\tcbuselibrary{listings,skins,breakable}

\newtcolorbox{promptbox}[2][]{enhanced,
  breakable,
  colback=gray!5,
  colframe=gray!60,
  fonttitle=\bfseries,
  coltitle=black,
  title={Prompt~\thetcbcounter: #2},
  label={#1},
  boxrule=0.5pt,
  arc=2pt,
  left=6pt,
  right=6pt,
  top=6pt,
  bottom=6pt,
}

\title{\reasoningvla{}: Bridging Reasoning and Action Prediction for Generalizable Autonomous Driving in the Long Tail}

\author{NVIDIA\footnote{A detailed list of contributors and acknowledgments can be found in~\cref{sec::contributors} of this paper.

\hspace{0.15cm} Following the release of NVIDIA Alpamayo at CES 2026~\citep{alpamayo2026ces}, Alpamayo-R1 is also referred to as Alpamayo 1.}}

\begin{abstract}
End-to-end architectures trained via imitation learning have advanced autonomous driving by scaling model size and data, yet performance remains brittle in safety-critical long-tail scenarios where supervision is sparse and causal understanding is limited. We introduce \reasoningvla (\reasoningvlashort), a vision–language–action model (VLA) that integrates \datafullname reasoning with trajectory planning for complex driving scenarios. 
Our approach features three key innovations: 
(1) the \datafullname (\datashortnamenosp) dataset, built through a hybrid auto-labeling and human-in-the-loop pipeline producing decision-grounded, causally linked reasoning traces aligned with driving behaviors;
(2) a modular VLA architecture combining Cosmos-Reason, a vision-language model pre-trained for Physical AI, with a diffusion-based trajectory decoder that generates dynamically feasible trajectories in real time;
(3) a multi-stage training strategy using supervised fine-tuning to elicit reasoning and reinforcement learning (RL) to enforce reasoning-action consistency and optimize reasoning quality.
\reasoningvlashort achieves up to a 12\% improvement in planning accuracy on challenging cases compared to a trajectory-only baseline, with a 35\% reduction in close encounter rate in closed-loop simulation. RL post-training improves reasoning quality by 45\% and reasoning-action consistency by 37\%. Model scaling from 0.5B to 7B parameters shows consistent improvements. On-vehicle road tests confirm real-time performance (99 ms latency) and successful urban deployment. By bridging interpretable reasoning with precise control, \reasoningvlashort demonstrates a practical path towards Level 4 autonomous driving. Model weights are available at \url{https://huggingface.co/nvidia/Alpamayo-R1-10B} with inference code at \url{https://github.com/NVlabs/alpamayo}.

\end{abstract}

\begin{document}

\maketitle

\abscontent
\section{Introduction}
\label{sec::intro}

The evolution of autonomous driving systems has witnessed a paradigm shift from traditional modular architectures \citep{urmson2008boss, paden2016survey, zhang2018apolloem, lefevre2014survey} to end-to-end (E2E) driving frameworks \citep{bojarski2016nvidia, hu2023uniad, gu2023vad, weng2024paradrive,wu2025alpamayo}, a transition increasingly embraced by industry. 
In contrast to modular designs that explicitly separate perception, prediction, and planning with hand-crafted intermediate representations, E2E approaches map raw sensor inputs directly to vehicle motion through jointly trained neural networks. This unified formulation eliminates manually engineered interfaces, enabling joint optimization and data-driven policy learning at scale. Recent advances in transformer-based architectures, coupled with large-scale driving datasets have further improved the overall performance and generalization of the E2E driving paradigm.
Despite these successes, current E2E approaches remain fragile in handling long-tail and safety-critical situations, where sparse supervision and the need for high-level reasoning pose significant challenges.. Consequently, a significant gap persists between the capabilities of existing E2E models and the requirements for achieving robust Level-4 autonomy with driving-specific reasoning capabilities.

Recent advances in large language models (LLMs)~\citep{achiam2023gpt,comanici2025gemini} offer a promising direction to address this \textit{reasoning} gap. LLMs have transformed artificial intelligence, 
with scaling laws~\citep{kaplan2020scaling} demonstrating that model performance improves predictably as compute and data increase. 
Beyond training-time scaling, recent frontier models such as OpenAI's o1~\citep{openaio1}, DeepSeek-R1~\citep{deepseekai2025deepseekr1}, and similar systems have introduced a new paradigm: \emph{inference-time reasoning}. 
Unlike traditional single-step answer generation, these models generate intermediate reasoning traces, denoted \textit{chains of thought}~\citep{wei2022chain},
that mimic human problem-solving strategies. 
This shift makes inference time a tunable resource: allocating more compute to deliberative reasoning often yields more accurate, robust, and verifiable decisions~\citep{yao2023tree}. This reasoning capability is particularly important for autonomous driving, where decision-making is inherently uncertain and safety-critical. Text-based reasoning further enables models to explore alternative outcomes in language space before committing to actions, 
offering several key advantages:
\begin{enumerate}[label=(\arabic*)]
\item \textit{improved safety} through explicit counterfactual reasoning and the potential for runtime safety cross-checks and monitoring;
\item \textit{better interpretability} via human-readable decision rationales; 
\item \textit{richer training signals} that can be used as verifiable rewards to boost long-tail performance. 
\end{enumerate}

VLMs/VLAs have been widely applied to autonomous driving~\citep{mao2023gpt, mao2023language, hwang2024emma, zhou2025opendrivevla, renz2025simlingo}, 
however, most approaches either lack explicit reasoning~\citep{wu2025alpamayo, zhou2025autovla, jiang2025irl} 
or perform reasoning in a free-form, unstructured manner~\citep{luo2025adathinkdrive, yuan2025autodrive, rowe2025poutine}. 
Such approaches struggle to generalize beyond training distributions, 
especially in ambiguous or compositional long-tail scenarios where strong domain priors are essential. 
Moreover, treating autonomous vehicle (AV) reasoning as a pure natural language processing (NLP) problem 
overlooks the rich structural knowledge inherent to driving: 
lane geometry, traffic rules, map priors, agent interactions, and dynamic constraints. 

We argue that effective reasoning for autonomous driving must be \emph{causally grounded} and \emph{structurally aligned} with the task of driving. 
Instead of generating verbose, unstructured narratives, 
reasoning traces should explicitly link observed scene evidence to concrete driving decisions through causal chains, 
and these decisions should directly condition or control low-level trajectory generation. 
The above design principle ensures that reasoning is not only an interpretability-enhancing addition, 
but rather a functional component that improves both training efficiency and closed-loop driving performance, 
particularly in safety-critical long-tail events.

In this work, we introduce \textbf{\reasoningvla}, a VLA that extends the vision-action (VA) model Alpamayo-VA~\citep{wu2025alpamayo} with structured reasoning capabilities, bridging reasoning and action prediction for generalizable autonomous driving. 
It addresses the challenges stated above through three key innovations:
\begin{enumerate}
    \item We develop a structured \textbf{\datafullname (\datashortnamenosp)} labeling framework that produces decision-grounded, causally-linked reasoning traces aligned with driving scenarios, supported by a hybrid human-in-the-loop and auto-labeling pipeline for scalable high-quality data generation.
    \item We employ a \textbf{diffusion-based action-expert trajectory decoder} built on flow matching~\citep{lipman2023flow, driess2025knowledge} to efficiently generate continuous, multi-modal trajectory plans that align with the language reasoning outputs while meeting real-time inference requirements.
    \item We adopt a \textbf{multi-stage training strategy} that builds upon the Cosmos-Reason VLM backbone, injects action modality for trajectory prediction, elicits reasoning via supervised fine-tuning on \datashortname data, and employs reinforcement learning (RL) to boost the reasoning quality, reasoning-action consistency and trajectory quality.
\end{enumerate}
Through extensive open-loop and closed-loop (simulation and onboard) evaluations, we demonstrate that \reasoningvlashort achieves substantial improvements over end-to-end baselines, with the largest gains in rare, safety-critical scenarios, while maintaining real-time inference performance (99ms end-to-end latency).

In the following sections, we present the detailed components of our framework. \cref{sec::related} reviews related work. \cref{sec::model} presents the model architecture and key design choices. \cref{sec::data_labeling} describes the proposed hybrid labeling pipeline and the resulting \datashortname dataset, specifically developed for reasoning-based VLA tasks in autonomous driving. \cref{sec::training} outlines our multi-stage training strategy, where each stage progressively enhances the model’s capabilities from improving general visual-language understanding in the AV domain, to generating action modalities, to strengthening reasoning ability and output alignment. Finally, \cref{sec::experiment} reports extensive evaluation results, demonstrating the effectiveness of our approach in both open-loop and closed-loop environments.

\section{Related Work}
\label{sec::related}
Our work builds upon recent advances in VLMs for autonomous driving, reasoning-augmented action models, and post-training alignment techniques. We organize our review around four key areas. First, we discuss the evolution from general-purpose VLMs~\citep{hwang2024emma, xu2024vlm} to action-oriented VLAs~\citep{zhou2025opendrivevla, renz2025simlingo} in autonomous driving (\cref{sec::related::vlas}), highlighting the shift toward embodied action prediction. Second, we examine reasoning VLAs (\cref{sec::related::reasoning_vlas}) that incorporate explicit chain-of-thought processes~\citep{wei2022chain, luo2025adathinkdrive, rowe2025poutine} for interpretable decision-making. Third, we review post-training alignment methods (\cref{sec::related::posttraining_alignment}), particularly RL from human feedback (RLHF)~\citep{christiano2017deep} and RL with verifiable rewards (RLVR)~\citep{deepseekai2025deepseekr1}, which form the foundation of our reasoning alignment approach. Finally, we review vision-language datasets in autonomous driving (\cref{sec::related::visual_language_datasets}), identifying key limitations in existing reasoning datasets~\citep{sima2024drivelm, nie2024reason2drive} that motivate our data construction methodology.

\subsection{VLMs and VLAs in Autonomous Driving}
\label{sec::related::vlas}
Early work explored leveraging LLMs' general knowledge for driving. Drive-GPT~\citep{mao2023gpt}, Wolf~\citep{li2024wolf}, and AgentDriver~\citep{mao2023language} treat planning as text generation or language-based tool use, achieving competitive open-loop performance. Cube-LLM~\citep{cho2024language}, TOKEN~\citep{tian2024tokenize}, and EMMA~\citep{hwang2024emma} scale multimodal LLMs to multi-task scene understanding and trajectory prediction. VLM-AD~\citep{xu2024vlm} uses VLMs as training-time supervisors, while ReAL-AD~\citep{lu2025real} models hierarchical reasoning, and DiMA~\citep{hegde2025distilling} distills VLM knowledge into efficient, LLM-free planners.

A complementary line of work couples language with explicit action representation to create VLA models. OpenDriveVLA~\citep{zhou2025opendrivevla} autoregressively produces trajectory waypoints from structured vision-language tokens. AutoVLA~\citep{zhou2025autovla} unifies reasoning and action with adaptive ``think vs. act'' control. IRL-VLA~\citep{jiang2025irl} incorporates inverse RL for safety-efficiency balance, CoReVLA~\citep{fang2025corevla} targets long-tail scenarios, and SimLingo~\citep{renz2025simlingo} achieves state-of-the-art closed-loop results in Bench2Drive~\citep{jia2024bench2drive}. However, these approaches largely operate reactively without explicit reasoning, struggling to generalize beyond training distributions in ambiguous or long-horizon scenarios requiring counterfactual reasoning.

\subsection{Reasoning VLAs in Autonomous Driving}
\label{sec::related::reasoning_vlas}
Explicit reasoning methods such as chain-of-thought (CoT)~\citep{wei2022chain} and tree-of-thought (ToT)~\citep{yao2023tree} have demonstrated that intermediate reasoning traces can substantially improve performance in complex language tasks. In the domain of autonomous driving, many recent works on VLA adopt this insight by integrating structured reasoning into vision-to-action pipelines.
One line of work focuses on adaptive or efficient invocation of reasoning. For example, AdaThinkDrive~\citep{luo2025adathinkdrive} uses a fast-and-slow thinking mechanism, trained with RL, to invoke CoT only when needed, reducing inference overhead while maintaining performance. AutoDrive-R$^2$~\citep{yuan2025autodrive} builds an explicit CoT and self-reflection dataset (nuScenesR$^2$-6K), leveraging GRPO~\citep{shao2024deepseekmath} with physics-grounded rewards to refine reasoning-augmented trajectories while ensuring physical feasibility.

Other approaches explore diverse reasoning strategies: RIV-CoT~\citep{corbiere2025retrieval} augments CoT with retrieval, FutureSightDrive~\citep{zeng2025futuresightdrive} performs spatio-temporal reasoning, and CoT-Drive~\citep{liao2025cot} distills reasoning into lightweight models. ReCogDrive~\citep{li2025recogdrive}, ReasonPlan~\citep{liu2025reasonplan}, MTRDrive~\citep{luo2025mtrdrive}, Drive-R1~\citep{li2025drive}, AgentThink~\citep{qian2025agentthink}, DriveAgent~\citep{hou2025driveagent}, and DSDrive~\citep{liu2025dsdrive} combine memory, tool invocation, multi-agent reasoning, or compression. Notably, Poutine~\citep{rowe2025poutine} topped the 2025 Waymo Vision-Based End-to-End Driving Challenge, demonstrating that reasoning-enhanced VLAs with RL finetuning excel in long-tail scenarios. This work demonstrates that reasoning serves as a functional core of driving decisions, with trade-offs among interpretability, runtime cost, and performance. However, most existing approaches rely on free-form reasoning that lacks explicit causal grounding and consistency between reasoning and actions. In contrast, our work introduces a structured \datashortname framework that ties reasoning to concrete driving decisions, and employs post-training RL to simultaneously optimize reasoning quality, reasoning-action consistency, and trajectory safety.

\subsection{Post-training Alignment}
\label{sec::related::posttraining_alignment}
Generative models (e.g., LLMs and text-to-image generators) are predominantly trained with an imitative objective, such as next-token prediction. While this objective enables efficient learning from Internet-scale data, it remains only a proxy for the true training goal: optimizing for the expert's internal reward function that motivated the demonstrated behavior. Consequently, generative models may deviate from end-user intent and, in some cases, exhibit safety-critical failures, such as producing harmful text outputs~\citep{murule}, unsafe visual generations~\citep{lee2023aligning}, or hazardous robot motions \citep{lu2023imitation}.
To mitigate such misalignment, post-training alignment—particularly through RLHF \citep{christiano2017deep} has emerged as a central strategy for aligning generative models with human preferences. For reasoning models specifically, DeepSeek-R1~\citep{deepseekai2025deepseekr1} employs Group Relative Policy Optimization (GRPO)~\citep{shao2024deepseekmath} to directly improve reasoning quality by rewarding verifiable solutions rather than intermediate token likelihood, while OpenAI o1~\citep{openaio1} similarly demonstrates that outcome-based RL substantially enhances chain-of-thought (CoT) quality. In the embodied AI domain, these alignment techniques have been extended to VLAs to generate actions that better reflect human intent across diverse embodiments, including autonomous driving~\citep{tian2025direct} and assistive robots~\citep{tian2024maximizing, zhang2025rewind}.
While these methods focus on improving action outcomes, our work addresses a complementary dimension: improving the reasoning process itself and ensuring that the model’s internal decision rationale remains causally consistent and contextually grounded in the context of safety-critical autonomous driving.

\subsection{Vision-Language Datasets for Autonomous Driving} 
\label{sec::related::visual_language_datasets}
Building upon the open-source {nuScenes}~\citep{caesar2020nuscenes} dataset, early work~\citep{qian2024nuscenes,wu2025nuprompt,tian2025nuscenes} primarily focuses on object-centric perception tasks, enabling VLMs to acquire general perception knowledge and improve object grounding in driving scenes. 
Beyond {nuScenes}, datasets such as {WOMD-reasoning}~\citep{li2024womd} and {DriveQA}~\citep{wei2025driveqa} extend vision-language annotations to large-scale motion datasets such as the Waymo Open Motion Dataset~\citep{waymo_open_motion_dataset} and the CARLA simulator~\citep{dosovitskiy2017carla}, focusing on describing interactions between agents, traffic rules, and right of way principles. 
While these datasets serve as valuable resources for VLM pre-training, their language annotations are not explicitly linked to the ego-vehicle's actions. As a result, they provide limited supervision for \textit{planning-oriented} reasoning, a key capability required by VLAs. To bridge this gap, prior work has focused on constructing language datasets tailored for motion planning. For instance, {Drama}~\citep{malla2023drama} annotates important objects that may influence the ego vehicle's behavior. Subsequent works such as {DriveAction}~\citep{hao2025driveaction} and {DriveBench}~\citep{xie2025vlms} develop comprehensive QA pairs for VLA training, emphasizing not only the identification of critical objects for planning, but also covering QA pairs for motion prediction, traffic signs, road markings, navigation following, etc. 

Motivated by the development of reasoning VLAs, recent research has shifted from general VLA datasets to reasoning-oriented ones, where explicit explanations are provided for the ego vehicle's actions. 
As an early effort, {BDD-X}~\citep{kim2018textual} provides a small set of human-written explanations describing driver behaviors. With the significant advancement of LLMs/VLMs, subsequent works such as DriveGPT4~\citep{xu2024drivegpt4}, CoVLA~\citep{arai2025covla}, and LingoQA~\citep{marcu2024lingoqa} introduce automated or human-in-the-loop pipelines to enrich the linguistic expressiveness of reasoning data. 
To capture the full reasoning process across perception, prediction and planning, DriveCoT~\citep{wang2024drivecot}, {Nuinstruct}~\citep{ding2024holistic}, Reason2drive~\citep{nie2024reason2drive}, DriveLM~\citep{sima2024drivelm}, DriveLMM-o1~\citep{ishaq2025drivelmm}, and Senna~\citep{jiang2024senna} develop explicit chain-based reasoning pipelines for data construction. 
In parallel, {Impromptu VLA}~\citep{chi2025impromptu} focuses on curating reasoning data in unstructured road scenarios. 
However, these datasets still exhibit key limitations in enforcing the causal relationship between observations and actions in their reasoning traces. For example, free-form reasoning traces tend to use vague descriptions such as \textit{``the ego vehicle should be cautious and watch out for ...''} rather than specifying actionable driving decisions. Additionally, many reasoning traces contain superficial causal factors such as \textit{``sunny weather'', ``wide roads'', ``... due to traffic rules''}, or introduce causal confusion by exposing the entire video clip in the labeling process and referencing future events that are not observable. These issues underscore the need for a dataset with explicit, decision-grounded, and causally linked reasoning traces, motivating our proposed \datashortname data pipeline.

\section{Building a Reasoning VLA Architecture}
\label{sec::model}

\begin{figure}[t]
    \centering
    \includegraphics[width=\linewidth]{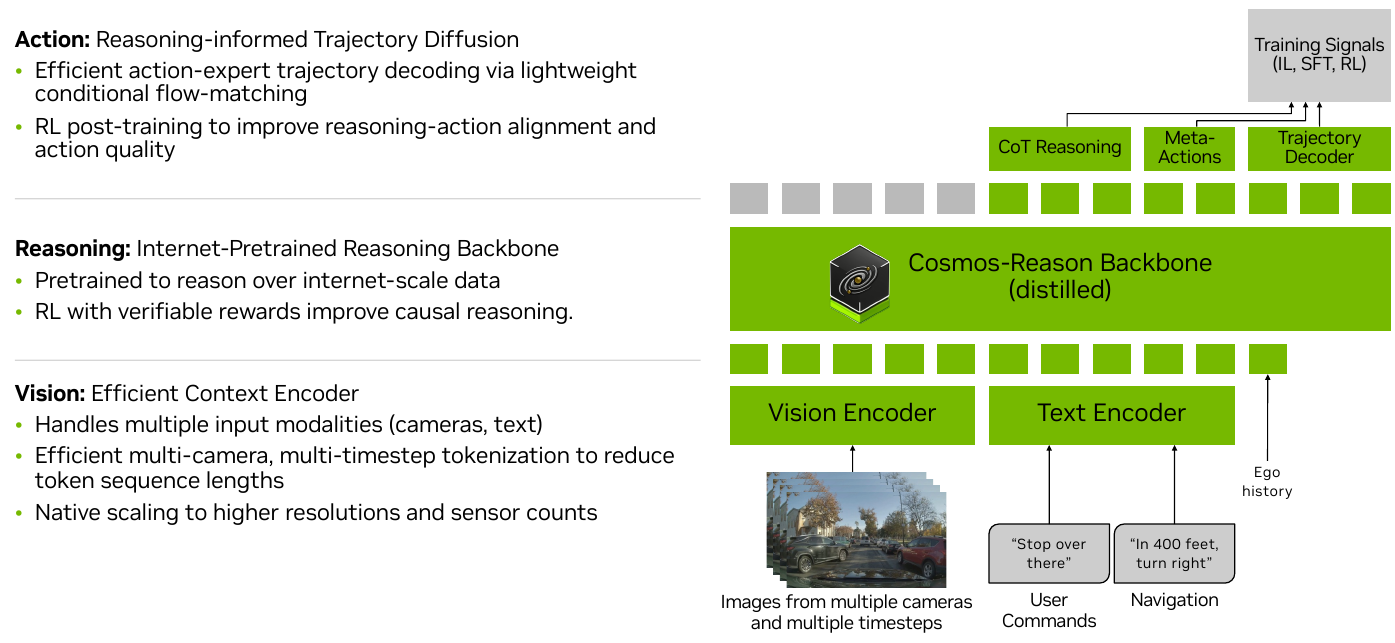}
    \vspace{-0.6cm}
    \caption{Overview of \reasoningvla architecture. Multi-camera images and egomotion are processed by a vision encoder to produce visual tokens, which are fed into a VLM backbone (Cosmos-Reason) along with textual inputs. The model autoregressively generates chain-of-thought reasoning and discrete trajectory tokens. At inference, an action-expert decoder using flow matching converts the discrete trajectory tokens into continuous, kinematically feasible waypoints conditioned on the reasoning output.}
    \label{fig:main_arch}
\end{figure}

Building an effective and reasoning-capable VLA for autonomous driving requires enabling several new capabilities beyond what general-purpose VLMs~\citep{achiam2023gpt,comanici2025gemini} currently offer. 
First, autonomous vehicles rely on \emph{multi-camera, multi-timestep observations} to achieve 360-degree situational awareness, yet standard VLMs typically process images or video frames \emph{independently} without explicit temporal or cross-view reasoning, leading to prohibitive token counts that preclude real-time inference when handling multi-camera inputs.
Second, driving decisions must be grounded in \emph{causally structured reasoning}~\citep{wei2022chain} rather than free-form narratives; the model must explain \emph{why} a maneuver is safe and legal based on observable evidence in the history window.
Third, the model must generate \emph{precise, multi-modal trajectory predictions} in real-time; autoregressively decoding waypoints as text tokens is inefficient and lacks the geometric and kinematic constraints essential for safe vehicle control~\citep{driess2025knowledge}.
Finally, to ensure safety in long-tail scenarios, reasoning traces must be \emph{aligned with executed actions}.

To address these challenges, we introduce \textbf{\reasoningvla (\reasoningvlashort)}, a modular VLA architecture that extends {Alpamayo-VA}~\citep{wu2025alpamayo} to integrate reasoning with action prediction for autonomous driving.
Our design philosophy emphasizes \emph{flexibility} and \emph{modularity}: the architecture can adopt any off-the-shelf VLM backbone while incorporating domain-specific components for efficient vision encoding and real-time action decoding.
This modularity enables us to leverage advances in vision-language pretraining~\citep{nvidia2025cosmosreason1physicalcommonsense, bai2025qwen2} while efficiently bridging high-level reasoning with low-level control for autonomous driving.

\textbf{Problem Formulation.} Given a sequence of past observations $\observation$ up to timestamp $T$ (omitted below), including multi-camera images $\observation_\text{image}$ and egomotion history $\observation_\text{egomotion}$, \reasoningvlashort is trained to perform reasoning, denoted as \reasoning, and to predict the future trajectory of the ego vehicle $\traj$.
We formulate this task as a sequential prediction problem,  
where the entire sequence is constructed as
\begin{equation}
    [\observation_\text{image}, \observation_\text{egomotion}, \reasoning, \traj],
    \label{eq:training_seq}
\end{equation}
with each component conditioned on all previous ones.
By default, the model is trained to predict the entire 6.4s-long future trajectory sequence
\begin{equation}
    \traj = \{(x^i, y^i, \theta_\text{yaw}^i)\}_{i=1}^{64},
    \label{eq:trajectory}
\end{equation}
where $(x^i, y^i, \theta_\text{yaw}^i)$ denotes the $i$-th future waypoint sampled at 10~Hz in the ego-vehicle's coordinate frame at time $T$.
As will be detailed in \cref{sec::action_decoding}, we adopt a control-based representation using unicycle dynamics with control inputs
\begin{equation}
    \dyntraj = \{(a^i, \kappa^i)\}_{i=1}^{64},
    \label{eq:dyn_trajectory}
\end{equation}
where $a^i$ and $\kappa^i$ denote the acceleration and curvature at timestep $i$, respectively.
Details of how $\traj$ is encoded and decoded are provided in \cref{sec::action_decoding,sec:traj_inject}.

\textbf{Overall Architecture.} \cref{fig:main_arch} presents the end-to-end architecture of \reasoningvlashort. The system processes multi-camera, multi-timestep observations as visual inputs, optionally augmented with textual inputs such as user commands and high-level navigation instructions. All inputs, including historical ego-motion data, are tokenized into a unified sequence of multimodal tokens following a predefined order. These tokens are then processed by the Cosmos-Reason~\citep{nvidia2025cosmosreason1physicalcommonsense} backbone, which produces output tokens representing reasoning traces, meta-actions, and predicted future trajectories. The model is trained in multiple stages, combining supervised fine-tuning (SFT) and RL, as will be described in \cref{sec::training}.

\subsection{VLM Backbone: Cosmos-Reason}
We adopt Cosmos-Reason~\citep{nvidia2025cosmosreason1physicalcommonsense} as the VLM backbone for \reasoningvla. Cosmos-Reason is a VLM specifically designed for Physical AI applications, post-trained on 3.7M Visual Question Answering (VQA) samples to develop physical common sense and embodied reasoning capabilities. The model incorporates 24.7K curated video VQA samples focused on driving scenarios, including scene descriptions, driving difficulty annotations, and reasoning traces distilled from DeepSeek-R1~\citep{deepseekai2025deepseekr1} to predict the next action.

\textbf{Domain-Specific Supervised Fine-Tuning.} To further enhance Cosmos-Reason for autonomous driving deployment, we curate supplementary datasets spanning multiple Physical AI domains, including autonomous driving, robotics, healthcare, smart cities, manufacturing, retail, and logistics. This broad Physical AI pre-training enables the model to develop general physical common sense and embodied reasoning capabilities that transfer to driving scenarios. For autonomous driving specifically, we augment the training data with 100K new samples that include annotations for critical objects in the environment and reasoning for the next action.

\mypara{Driving-Focused Data Curation.} We develop complementary labeling approaches to balance quality and scale for autonomous driving:

\begin{itemize}
    \item \textit{Human-labeled data} includes comprehensive annotations covering the operational design domain (weather, lighting, road conditions), traffic regulations (traffic lights, signs), ego behaviors (interactive and non-interactive meta-actions), critical objects influencing ego behavior, and causal reasoning behind observed maneuvers. These labels improve the model's understanding and reasoning in complex driving scenarios.
    \item \textit{Automatically labeled data} focuses on ego behavior reasoning and prediction, generated by prompting a teacher VLM (e.g., Qwen3-VL~\citep{qwen2025qwen3vl}) with driving-specific priors that encode longitudinal, lateral, and lane-related meta-actions along with velocity information. This scalable approach strengthens the model's predictive reasoning capabilities.
\end{itemize}

\subsection{Domain-Specific Adaptations} 
While Cosmos-Reason provides a strong foundation, two critical gaps remain for real-world autonomous driving deployment: efficient vision encoding for multi-camera, multi-timestep inputs and precise trajectory decoding for real-time control. The following subsections detail our domain-specific components that address these challenges.

\subsubsection{Vision Encoding}
\label{sec::vision_encoding}

The main role of vision encoders within VLMs is to convert input images into streams of tokens for later processing by the LLM backbone. However, as VLAs target onboard deployment, a critical requirement of their vision encoders is to produce as few tokens as possible while preserving relevant semantic information from the environment. To achieve this, there have been a variety of vision tokenization approaches proposed that primarily differ in how much information is encoded per inference step (i.e., how many images are compressed into how many tokens), as well as their associated architectural choices.

In this section, we discuss the different vision encoders that \reasoningvlashort{} can use as well as their tradeoffs, in addition to avenues for further token count compression towards enabling real-time onboard inference with larger backbone sizes.

\textbf{Single-Image Tokenization.}
Many vision tokenizers primarily focus on representing single images and either employ autoencoding architectures~\citep{SohnLeeEtAl2015,OordVinyalsEtAl2018,esser2021taming} or directly encode patches of pixels~\citep{dosovitskiy2020image}. VLMs adopt the latter primarily and employ Vision Transformers (ViTs)~\citep{dosovitskiy2020image} to partition images into patches that are encoded to form a 1D token sequence. We denote this paradigm as \textit{single-image tokenization}, where a model encodes each input frame into a set of tokens.

\reasoningvlashort{}'s default tokenizer (and the one used for all subsequent experiments) leverages this paradigm, employing the base VLM's vision encoder (e.g., \citet{zhai2023siglip,tschannen2025siglip2}) to encode a $W \times H$ px input image into patch features $\mathbf{f} \in \mathbb{R}^{W/14 \times H/14 \times D}$ which are then $2\times$ bilinearly downsampled to $\mathbf{f}' \in \mathbb{R}^{W/28 \times H/28 \times D}$ features per image. As an example, with $W = 448, H = 280$ this process produces 160 tokens per image.

\textbf{Multi-Camera Tokenization.}
While single-image tokenization is simple to implement, it produces token counts that scale linearly with image resolution and the number of cameras~\citep{wang2025patchscaling}.
To obtain a 360-degree view of their surroundings, AVs often use 6 to 10 cameras, the patch-based tokenization of which would yield thousands of tokens per timestep and preclude real-time inference. Accordingly, \reasoningvlashort{} also supports the use of a new line of efficient multi-camera tokenizers that encode images from multiple cameras into an intermediate representation before tokenizing that representation.

Specifically, \reasoningvlashort{} can additionally use the efficient multi-camera tokenizer proposed in~\citet{ivanovic2025efficient}, which leverages triplanes as a 3D inductive bias, to simultaneously represent multiple camera images in an efficient manner.
Crucially, since the triplane sizes are fixed, the input number of cameras and their resolution are decoupled from the resulting number of tokens. 
Formally, for a triplane with grid sizes $S_x, S_y, S_z$ and downstream patchification values of $p_x, p_y, p_z$, the number of tokens produced by the tokenizer is
\begin{equation}
    \underbrace{\left(\frac{S_x-p_x}{p_x}+1\right)\left(\frac{S_y-p_y}{p_y}+1\right)}_{\text{\# of patches in the } xy \text{ plane}}+\underbrace{\left(\frac{S_x-p_x}{p_x}+1\right)\left(\frac{S_z-p_z}{p_z}+1\right)}_{\text{\# of patches in the } xz \text{ plane}}+\underbrace{\left(\frac{S_y-p_y}{p_y}+1\right)\left(\frac{S_z-p_z}{p_z}+1\right)}_{\text{\# of patches in the } yz \text{ plane}}.
\end{equation}
As an example, for $S_x = S_y = 96, S_z = 48$, and $p_x = p_y = p_z = 8$, only 288 tokens are needed to represent one timestep of observations, irrespective of the number of cameras or their resolution. For a 7-camera vehicle setup, this equates to approximately $41.1$ tokens per image ($3.9\times$ less than single-image tokenization). Further, as we will show in~\cref{sec::exp::visenc}, this efficiency is achieved without major compromises to end-to-end driving metrics.

\textbf{Multi-Camera Video Tokenization.} While the above already yields significant reductions in the number of tokens required to represent sensor observations, there are still two fundamental areas where additional efficiency can be achieved:
\begin{enumerate}[label=(\arabic*)]
    \item accounting for temporal information (e.g., there can be redundancy in information across frames);
    \item removing the potential performance ceiling that comes with using a structured feature representation.
\end{enumerate}
Accordingly, \reasoningvlashort{} also supports using multi-camera video tokenizers that directly encode entire sequences of camera observations from multiple timesteps. One example is Flex~\citep{yang2025flex}, which compresses a set of image tokens from multiple cameras and timesteps via full self-attention layers and a fixed set of query vectors, providing an explicit mechanism to control the magnitude of the information bottleneck. As will be shown in~\cref{sec::exp::visenc}, this approach can achieve an up to $20\times$ token compression rate (compared to single-image tokenization) while maintaining or even improving downstream driving metrics.

\textbf{Additional Avenues for Token Compression.}
Beyond the tokenization strategies described above, several complementary approaches can further reduce token counts. Post-training token pruning techniques, exemplified by SparseVILA~\citep{khaki2025sparsevila}, dynamically identify and remove redundant tokens during inference without retraining, offering a practical path to reduce computational costs on models already trained. 
These methods represent promising directions for further scaling \reasoningvlashort{} to even larger backbones while maintaining real-time performance constraints.

\subsubsection{Trajectory Decoding}
\label{sec::action_decoding}

\definecolor{lightyellow}{RGB}{245, 210, 60}  %
\definecolor{skyblue}{RGB}{40, 100, 220}      %
\definecolor{softred}{RGB}{220, 60, 50}       %

\begin{figure}[t!]
    \centering
    \includegraphics[width=\linewidth, trim={20 100 20 100}, clip]{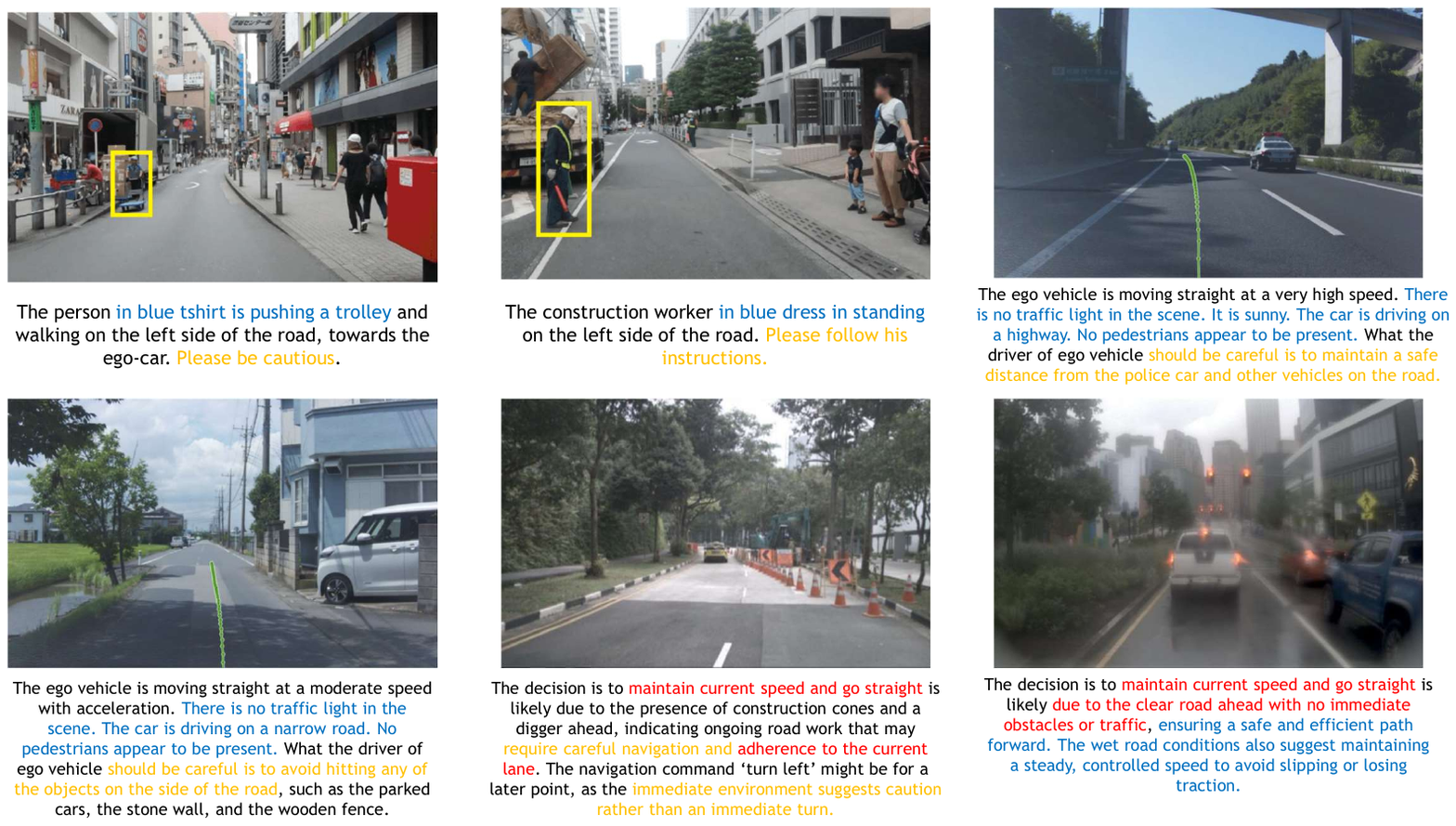}
    \vspace{-0.4cm}
    \caption{Examples of reasoning traces exhibiting common issues in existing datasets~\citep{malla2023drama, chi2025impromptu, arai2025covla}. Text highlighted in \textcolor{lightyellow}{yellow} indicates vague behavior descriptions that fail to specify concrete driving decisions correlated with the trajectories. Text highlighted in \textcolor{skyblue}{blue} denotes superficial reasoning, such as contextual observations that do not directly inform the ego vehicle’s decision. \textcolor{softred}{Red} highlights indicate incorrect or causally inconsistent reasoning that contradicts the actual behavior of the ego vehicle.}
    \label{fig:cot_failure}
\end{figure}

To extend the capability of a VLM to operate effectively in the physical world, it is essential to incorporate \textit{physical actions}, corresponding to future driving trajectories in the autonomous driving context, into the training of the VLA. However, embodiment introduces unique challenges in action decoding: 
\begin{enumerate}[label=(\arabic*)]
    \item the action representation must be accurate, preserving both fidelity and multi-modality;
    \item the decoding process must be fast enough to support real-time inference;
    \item the decoding mechanism should integrate seamlessly into the VLA training pipeline.
\end{enumerate}

Initially, we found that training the model in raw position (i.e., $x, y$) waypoint space is susceptible to sensor noise, which often degrades model convergence. 
Moreover, the downstream low-level vehicle controllers typically smooth trajectory outputs to ensure consistent and stable execution on-vehicle.
Thus, instead of directly learning $\traj$ in the raw position waypoint space, we adopt an action representation governed by unicycle dynamics that leads to better closed-loop performance. 
Specifically, we employ the following unicycle dynamics with control input $\dyntraj = \{(a^i, \kappa^i)\}_{i=1}^{64}$~\citep{lynch2017modern} and apply Euler discretization:
\begin{equation}
\label{eqn::dynamics}
\mathbf{x}^{i+1}
= \begin{pmatrix}
x^{i+1}\\
y^{i+1}\\
\theta^{i+1}\\
v^{i+1}\\
\end{pmatrix}
= \begin{pmatrix}
x^i + \dfrac{\Delta T}{2}\!\left( v^i \cos\theta^i + v^{i+1}\cos\theta^{i+1} \right)\\[6pt]
y^i + \dfrac{\Delta T}{2}\!\left( v^i \sin\theta^i + v^{i+1}\sin\theta^{i+1} \right)\\[6pt]
\theta^i + \Delta T\,\kappa^i v^i + \dfrac{\Delta T^{2}}{2}\,\kappa^i a^i\\[6pt]
v^i + \Delta T\,a^i
\end{pmatrix},
\end{equation}
where $\Delta T = 0.1\text{s}$ in our setup, $x$ and $y$ denote positional waypoints in the bird’s-eye-view (BEV) plane, $\theta$ represents the yaw angle, $v$ the velocity, $\kappa$ the curvature, and $a$ the acceleration.
During training, the ground-truth control sequence $\dyntraj$ is derived from $\traj$ through a least-squares formulation with Tikhonov regularization to attenuate high-frequency noise. 
The model is trained to predict the control sequence $\dyntraj$ and, during inference, we apply \cref{eqn::dynamics} to map it to $\traj$.

Furthermore, to enable \reasoningvlashort to understand and generate trajectories, we encode $\traj$ either as discrete tokens or continuous embeddings. In the discrete representation, we uniformly quantize each continuous value in $\dyntraj$ within a predefined range into equally spaced bins and represent the resulting indices as special tokens. For the continuous representation, we map $\dyntraj$ into \reasoningvlashort's embedding space using sinusoidal positional encoding followed by an MLP projection. 
Specifically, we adopt a strategy inspired by $\pi_{0.5}$-KI~\citep{driess2025knowledge}, combining \textit{discrete} trajectory tokens learned within the VLM with an action-expert that decodes the same trajectories into \textit{continuous} representations using a flow matching framework~\citep{lipman2023flow}. This framework facilitates streamlined VLM training, accelerates trajectory decoding, and achieves better closed-loop performance.
Training details of the action modality injection are provided in \cref{sec:traj_inject}.

\textbf{Summary.}
This section further detailed the two principal design dimensions (vision encoding and action decoding) through which VLMs can be systematically adapted into AV policy VLAs. In subsequent sections, we detail the construction of the data pipeline and the formulation of the training strategy, which together endow the model with enhanced reasoning and alignment capabilities, thereby improving its robustness in handling long-tail events.

\begin{figure}[t!]
    \centering
    \includegraphics[width=\linewidth]{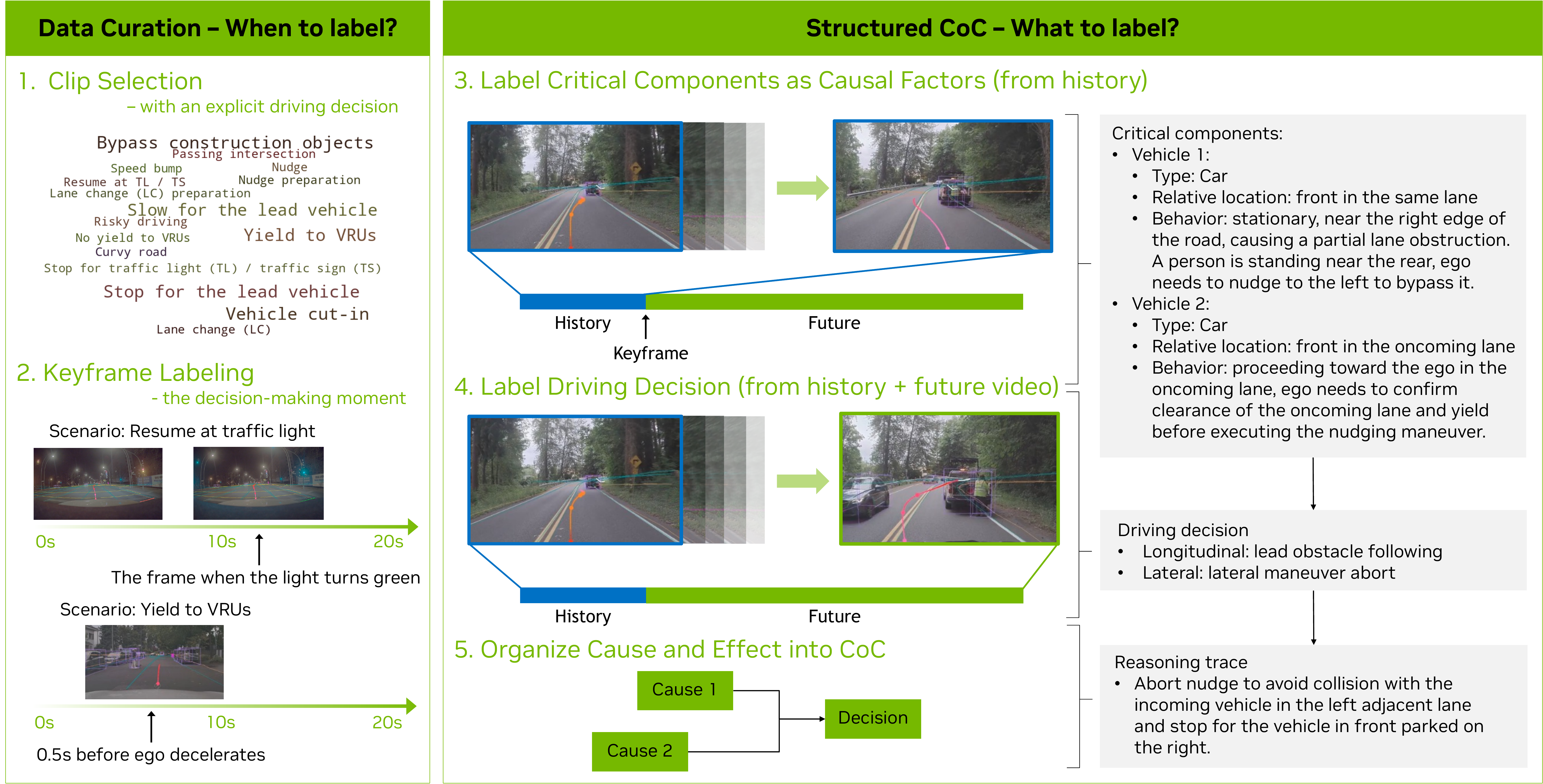}
    \vspace{-0.4cm}
    \caption{Overview of the proposed structured \datashortname labeling pipeline, composed of five steps: (1) \textbf{Clip Selection}, where clips containing explicit driving decisions are selected, filtering out low-signal clips that offer limited causal information; (2) \textbf{Keyframe Labeling}, where the decision-making moment within each video clip is identified, minimizing potential causal confusion; (3-5) \textbf{Structured \datashortname Labeling}, to construct the final \datashortname and further mitigate causal confusion, we first annotate critical components from the observation while avoiding referring to causal factors in future frames, and then label the corresponding driving decision. We then compose a reasoning trace from driving decisions and causal factors in natural language.
    } 
    \label{fig:labeling_pipeline}
\end{figure}

\newcommand{\timebuffernosp}{0.5\,seconds}
\newcommand{\timebuffer}{\timebuffernosp~}

\definecolor{lightyellow}{RGB}{245, 210, 60}  %
\definecolor{skyblue}{RGB}{40, 100, 220}      %
\definecolor{softred}{RGB}{220, 60, 50}       %

\section{\datafullname Dataset: Learning Causally Grounded Reasoning VLAs}
\label{sec::data_labeling}

To enable reasoning VLA models to explain the causes of driving actions and to improve their trajectory-level performance, reasoning data must be closely correlated with the ego trajectory. However, existing CoT reasoning datasets in the AV community often exhibit several limitations, as shown in \cref{fig:cot_failure}:
\begin{enumerate}[label=(\arabic*)]
\item \textit{Vague behavior descriptions}: free-form CoT annotations may fail to specify concrete driving actions or may choose words that weakly correlate with ego trajectories;
\item \textit{Superficial reasoning}: some reasoning traces primarily describe contextual observations or hypothetical factors that lack a direct causal link to the ego vehicle's behavior, providing limited benefit for improving post-training driving performance;
\item \textit{Causal confusion}: reasoning traces may include causal factors that occur in future time windows, which are not observable to the model during training. This arises because the labeling process often exposes the entire video without distinguishing between historical and future segments.
\end{enumerate}

To address these gaps, we introduce a labeling framework that enforces an explicit causal structure in the reasoning traces. We first define a comprehensive set of high-level driving decisions that directly correspond to low-level ego trajectories. Each reasoning trace is associated with an explicit driving decision and includes only the causal factors that motivate that driving decision. By carefully selecting keyframes to split historical and future video segments, we ensure that all causal factors originate within the observable history window, thereby preventing causal confusion. This design ensures that every reasoning trace is both \textit{decision-grounded} and \textit{causally linked}, capturing concise and interpretable cause–and–effect relationships rather than verbose descriptive narratives. The resulting dataset, termed the \textbf{\datafullname (\datashortnamenosp) dataset}, provides clear supervision for learning decision causality, enabling reasoning VLAs to efficiently reason about the causes of specific driving actions during onboard inference. An overview of our labeling pipeline is shown in \cref{fig:labeling_pipeline}.

\subsection{Structured Chain of Causation} 
\label{sec:structured_coc}

To facilitate efficient annotation, our labeling framework decomposes each data sample into three structured components: the driving decision, the causal factors (critical components), and the composed \datashortname trace. Consequently, each data instance constitutes a structured \datashortname sample encompassing these three components.

\begin{table}[t]
\centering
\small
\caption{Closed-set driving decisions (longitudinal and lateral) used to anchor reasoning traces to explicit control intent. Annotators select at most one decision per channel (or \emph{None}), ensuring decision-grounded supervision. 
Definitions emphasize operational intent and disambiguate visually or behaviorally similar maneuvers (e.g., \emph{Lead obstacle following} vs.\ \emph{Yield}, \emph{Lane change} vs.\ \emph{Merge / Split}). Each selected decision must be causally supported by evidence from the observed history window. LC denotes lane change.}
\vspace{-0.2cm}
\begin{tabularx}{\linewidth}{ll X}
\toprule
\textbf{Type} & \textbf{Driving decision} & \textbf{Definition} \\
\midrule
Longitudinal & Set speed tracking & Maintain or reach a target speed when unconstrained; excludes follow/yield/stop logic. \\
& Lead obstacle following & Maintain a safe time gap to the lead entity (closest in-path entity moves in the same traffic flow); excludes geometry-based slowing, gap-matching, and yielding to non-lead entity. \\
& Speed adaptation (road events) & Adjust speed for roadway features (curves, grades, bumps, ramps, roundabouts, turns); independent of a lead. \\
& Gap-searching (for LC/merge/zipper) & Adjust speed to match the target stream or create a usable gap to support a planned lateral maneuver. \\
& Acceleration for passing/overtaking & Increase speed to pass a slower lead with an associated lateral plan. \\
& Yield (agent right-of-way) & Slow/stop to concede priority to specific agents (pedestrians, cross-traffic, emergency vehicles, cut-ins). \\
& Stop for static constraints & Decelerate to—and hold at—control points (stop/yield lines, red light, school bus/rail rules); Sometimes a yield is necessary even when owning the right-of-way, to avoid a collision.  \\
\midrule
Lateral & Lane keeping \& centering & Maintain position within lane boundaries; minor in-lane offsets allowed; never cross lane lines. \\
& Merge / Split (facility change) & Transition between facilities (e.g., on-ramp $\leftrightarrow$ mainline, weave segments); not a same-road lane change. \\
& Out-of-lane nudge (straddle avoidance) & Brief, intentional lane-line crossing to increase clearance around a blockage/hazard; return to original lane; specify left/right. \\
& In-lane nudge & Temporary offset within the lane (no line crossing) to increase clearance around a blockage/hazard; specify left/right. \\
& Lane change (lateral push) & Full adjacent-lane transition with gap negotiation; specify left/right in reasoning trace. \\
& Pull-over / curb approach & Move toward edge/shoulder or a designated stop area (pickup, emergency stop, parking approach). \\
& Turn (intersection/roundabout/U-turn) & Planned path onto a different road segment with a significant heading change; specify left/right. \\
& Lateral maneuver abort & Cancel an ongoing lateral maneuver (nudge, lane change, merge/split, pull-over) and re-center when safe. \\
\bottomrule
\end{tabularx}
\label{tab:driving_decisions}
\end{table}

\mypara{Driving Decision.} To ensure our \datashortname data is \emph{decision-grounded}, we define a closed set of high-level driving decisions as in \cref{tab:driving_decisions}. Each clip is annotated with at most one longitudinal and one lateral decision (or \emph{None} for either channel), corresponding to the first action taken by the ego vehicle immediately after the critical reasoning moment. This standardized inventory directly aligns with low-level trajectories and eliminates free-form, vague descriptions of driving behavior, ensuring that every reasoning trace unambiguously specifies \emph{what} decision is taken. For linguistic consistency and diversity, the final \datashortname reasoning traces are constructed using a compact verb set aligned with these driving decisions.

\begin{table}[t]
\centering
\small
\caption{Categories and example attributes of \textit{critical components} that may serve as causal factors for driving decisions. Only those directly influencing the driving decision are labeled. Use a Low/High uncertainty tag when forecasting object behavior or when signals are partially occluded. The list is open-ended, allowing additional critical components to be added as needed.}
\vspace{-0.2cm}
\begin{tabular}{@{}p{0.17\linewidth} p{0.66\linewidth} p{0.1\linewidth}@{}}
\toprule
\textbf{Category} & \textbf{Example attributes to record (if decision-relevant)} & \textbf{Uncertainty} \\
\midrule
\textbf{Critical objects} & Type (veh./ped./cyclist/VRU), relative pose to ego (in-path, left/right, oncoming, crosswalk), motion (stopped, slowing, crossing, cut-in risk) & Low / High \\
\textbf{Traffic lights} & Current state (R/Y/G), arrow state, visibility/occlusion; presence of wait line & - \\
\textbf{Yield/Stop control} & Presence of signs, all-way vs two-way, stop/yield line location & - \\
\textbf{Road events} & Curvature/grade, speed bump, narrowing, roundabout, ramp/junction ahead & - \\
\textbf{Lane / lanelines} & Lane count, laneline type (dashed/solid), shoulder/bike lane, usable width & - \\
\textbf{Routing intent} & Target lane/turn (L/R/through), near-term split/merge, required lane for maneuver & - \\
\textbf{ODD constraints} & Weather/visibility, construction, emergency vehicles, school bus/rail rules & - \\
\bottomrule
\end{tabular}
\label{tab:critical_components}
\end{table}

\mypara{Critical Components.} In contrast to the closed-set driving decisions, causal factors are defined as an open-ended set, with categories and example attributes described in \cref{tab:critical_components}. This design allows human labelers or an auto-labeling pipeline to flexibly specify only the key elements that directly influence the driving decision, while maintaining a structured output.

\mypara{Composed CoC Traces.} Once the driving decision and critical components are identified, they are linguistically organized into a coherent \datashortname reasoning trace that captures the causal rationale behind the chosen decision. As a result, the structured \datashortname protocol enforces:
\begin{enumerate}[label=(\arabic*)]
\item \textit{decision grounding}: each reasoning trace is anchored to a single, explicit decision at the critical moment;
\item \textit{causal locality}: all evidence must originate from the observed history window;
\item \textit{annotation economy}: only decision-relevant factors are included.
\end{enumerate}

\subsection{Data Curation} %

Having defined the structured components of \datashortname (driving decisions, critical components, and composed \datashortname traces), the next step is to determine when these reasoning data should be labeled. Not every video clip warrants annotation; labeling is triggered only at moments where a clear causal link can be established between observable factors and the ego vehicle’s subsequent decision. Therefore, a key aspect of our data labeling framework is data curation, which involves identifying these critical reasoning moments.

\mypara{Clip Selection.} We choose clips that contain an explicit driving decision to label the \datashortname dataset, thereby avoiding low-signal clips that provide limited causal information. These clips are categorized into two types of scenarios: (1) \textit{Reactive} - where the ego vehicle must immediately adapt its behavior in response to a specific event, such as stopping for a lead vehicle or red light, or adjusting its lateral position to maintain clearance from a nearby obstacle or hazard; (2) \textit{Proactive} - where the ego vehicle is not required to react instantly but must actively assess and anticipate potential maneuver adjustments due to upcoming road events or obstacles. For example, the ego may receive a routing command to change lanes but lacks sufficient space in the target lane, requiring continuous gap searching and space assessment in preparation for the lane change maneuver. We employ rule-based methods to identify clips corresponding to each scenario and balance the number of clips per scenario to ensure dataset diversity. Detailed definitions of the scenarios are provided in \cref{tab:odd}.

\begin{table}[t]
\centering
\small
\renewcommand{\arraystretch}{1.2} %
\caption{Scenarios used in clip selection, along with keyframe and keyframe range definitions for \datashortname annotation. The goal is to identify critical reasoning moments within each selected clip, where a clear causal link can be established between observable factors and the ego vehicle’s subsequent decision.}
\vspace{-0.2cm}
\begin{tabular}{@{}p{1.4cm} p{3.6cm} p{10.6cm}@{}}
\toprule
\textbf{Type} & \textbf{Scenario name} & \textbf{Keyframe Definition (Reactive) / Keyframe Range (Proactive)} \\
\midrule
Reactive 
& Slow for the lead vehicle & \timebuffer before the ego decelerates behind a lead vehicle. \\
& Stop for the lead vehicle & Same as above. \\
& Stop for traffic light (TL) / traffic sign (TS) & Whichever occurs later: (1) \timebuffer before the ego begins to decelerate for a TL/TS; or (2) for a TL, the frame when it turns yellow/red. \\
& Resume at TL / TS & Whichever occurs later: (1) \timebuffer before the ego begins to accelerate from standstill due a TL/TS; or (2) for a TL, the frame when it turns green. \\
& Lane change (LC) & \timebuffer before the ego starts to move off-center of its original lane.  \\
& Yield to VRUs & \timebuffer before the ego begins to decelerate or nudge for a VRU. \\
& Vehicle cut-in & Whichever occurs first: (1) when the contender signals a LC into ego's lane; or (2) when the contender starts to move off-center of its original lane for the LC if no blinker signal is given. \\
& Speed bump & \timebuffer before the ego decelerates for the speed bump ahead. \\
& Nudge & \timebuffer before the ego moves away from the lane center to avoid or give space to an obstacle. \\
& Bypass construction objects & \timebuffer before the ego decelerates or nudges to construction objects or changes lane in response to construction objects modifying the lane. \\
& Risky driving & \timebuffer before the ego decelerates, nudges or moves backward for a risky event or obstacle, e.g., lane-weaving leading vehicle, parked vehicle backing out, or oncoming vehicle crossing into ego's lane. \\
\midrule
Proactive
& Curvy road & Start: whichever occurs first - (1) \timebuffer before the ego begins to decelerate for the curve; or (2) when the ego enters the curve at current speed. End: when the ego exits the curve. \\
& Lane change (LC) preparation & Start: ego receives a reason to perform a LC (e.g., route or passing a slow lead) but cannot do it immediately due to a blocked target lane.
End: Ego is ready to change lanes after gap searching or when traffic clears. \\
& Nudge preparation & Start: ego receives a reason to nudge for an obstacle, but cannot do it immediately due to traffic. End: Ego is ready to nudge once the traffic clears.  \\
& Passing intersection & Start: ego enters the intersection when the front bumper crosses the stop line or crosswalk boundary. End: ego fully exits the intersection area. \\
& No yield to VRUs & Start: when VRUs appear with the intention to cross but are not yet crossing because (1) ego has the right of way, or (2) VRUs intentionally yield to ego. End: when the VRUs are no longer visible.\\
\bottomrule
\end{tabular}
\renewcommand{\arraystretch}{1} %
\label{tab:odd}
\end{table}

\mypara{Keyframe Labeling.} Each raw clip contains 20 seconds of data and can generate multiple training samples, given the configuration of using a 2-second history to predict a 6-second future during both training and evaluation. Selecting keyframes for \datashortname annotation is therefore critical to maximizing the clarity of decision causality. For \textit{reactive} scenarios, a keyframe is typically chosen by applying a short temporal buffer (approximately \timebuffernosp) before the ego vehicle initiates a behavior change corresponding to a driving decision. At this keyframe, the ego vehicle has accumulated sufficient observations within the preceding 2-second history to justify the forthcoming action, effectively avoiding casual confusion. Because the keyframe is positioned immediately prior to the decision-making moment, we ensure that a concrete driving decision is associated with the data sample, enabling the annotation of decision-grounded \datashortname traces. For \textit{proactive} scenarios, we annotate a keyframe range: a time window during which the ego actively evaluates or prepares for a potential maneuver change. Detailed definitions of the keyframe or keyframe range for both reactive and proactive scenarios are provided in \cref{tab:odd}. \datashortname reasoning traces are annotated only for samples corresponding to the keyframe timestamp or the keyframes sampled from the keyframe range.

\subsection{Hybrid Labeling Procedure}

To ensure both quality and scalability, we develop a hybrid labeling procedure that combines human labeling and auto-labeling. While auto-labels are sufficient for generating large-scale training data for reasoning VLA models, high-quality and human-verified data, on the order of $\sim 10\%$ of the total, is essential for further SFT, auto-label evaluation, and model evaluation. Our proposed hybrid labeling approach balances efficiency and accuracy, supporting both large-scale training and reliable model assessment.

\subsubsection{Human Labeling} 
\label{subsec::human_labeling}

\mypara{Two-Stage Labeling Procedure.} Following the structured \datashortname described in \cref{sec:structured_coc}, human annotators are required to complete a two-stage procedure designed to produce concise and causally grounded \datashortname write-ups. 
\begin{enumerate}[leftmargin=*, itemsep=2pt]
\item \textbf{Stage I (0--2\,s):} identify \emph{critical components} from \cref{tab:critical_components} within the observed history window (within 2s before the keyframe). This step helps prevent causal confusion by ensuring that only evidence available prior to the decision-making moment is considered. These critical components may influence the driving decision annotated in the next stage.
\item \textbf{Stage II (0--8\,s):} (a) apply a safety exclusion filter to remove invalid data with illegal or unsafe driving behavior, (b) select the first post-keyframe driving decision for each channel (longitudinal and lateral; or \emph{None}), (c) write a \datashortname reasoning trace that references only the causal factors identified in Stage~I that lead to the driving decision, along with routing or regulatory signals when applicable.
\end{enumerate}
To enforce a clear separation between Stage I and Stage II and minimize causal leakage, we designed a labeling tool that explicitly distinguishes historical video segments (0-2\,s) from future segments (2-8\,s). This tool also provides visual aids, including ego-dynamics plots (speed, acceleration, steering angle, and turn signals), BEV visualizations overlaid with lane topology, and obstacle bounding boxes in order to help annotators achieve a more accurate understanding of the driving scene. 

\begin{figure}[t]
    \centering
    \includegraphics[width=\linewidth]{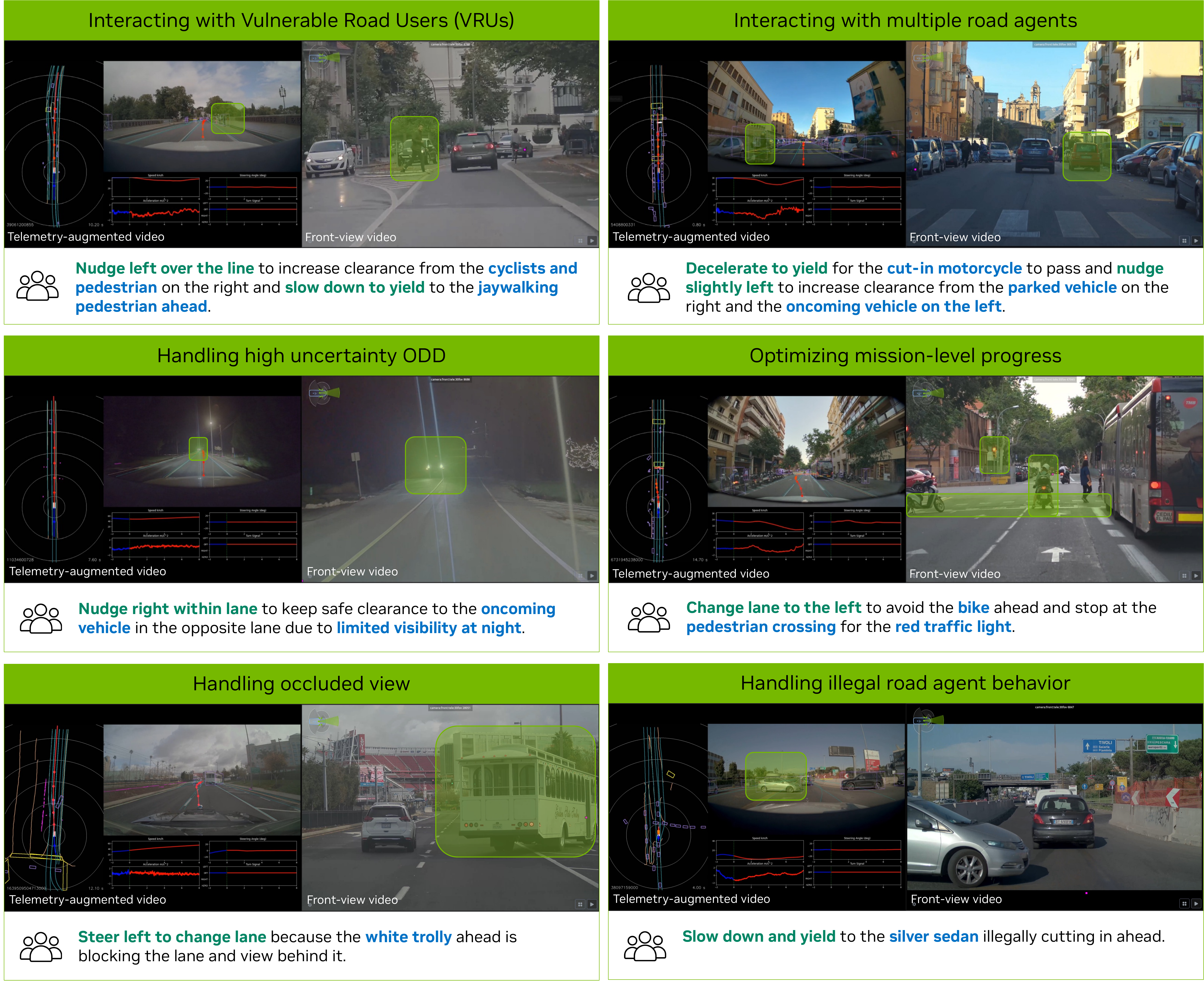}
    \vspace{-0.6cm}
    \caption{Examples of our labeled \datashortname reasoning traces, where \textcolor{ForestGreen}{\textbf{driving decisions}} and \textcolor{RoyalBlue}{\textbf{critical components}} are organized into \datashortname and highlighted correspondingly.}
    \label{fig:human_coc_examples}
\end{figure}

\mypara{Quality Assurance (QA).} To maximize annotation quality and reduce potential bias, we implement a rigorous QA process. Each labeled instance first undergoes a quality check performed by a different annotator. Moreover, $10\% - 20\%$ of labeled instances are selected, based on the performance of the assigned annotators, for an additional auditing process conducted by a dedicated team of experienced auditors. Both the quality check and auditing process follow the same QA guidelines, with key rules summarized in \cref{tab:human_qa}. This QA process ensures that the desiderata of \datashortname are rigorously enforced while preserving flexibility for natural language expression. As a result, we generate high-quality \datashortname reasoning traces across diverse driving scenarios, with representative examples shown in~\cref{fig:human_coc_examples}.

\begin{table}[t]
\centering
\small
\caption{Quality assurance (QA) checklist for quality check and auditing process. Key rules tie closely to the desiderata of \datashortnamenosp: decision grounding, causal locality, and annotation economy.}
\vspace{-0.2cm}
\begin{tabular}{p{4.1cm} p{9.4cm}}
\toprule
\textbf{Rule} & \textbf{Operational check} \\
\midrule
Causal coverage & Each selected decision references at least one Stage~I component; otherwise mark \emph{UNOBSERVED} with brief justification. \\
Causal correctness & Reasoning trace must logically explain the selected decision based on valid cause–and–effect relationships. Circular reasoning, misattributed causes, or missing necessary conditions are flagged for rework \\
Proximate cause & Prefer the immediate driver (e.g., stopped lead) over background conditions (e.g., red light when not first in queue). \\
Decision minimality & If no change in decision, label \emph{None}. \\
\bottomrule
\end{tabular}
\label{tab:human_qa}
\end{table}

\begin{table}[t]
\centering
\small
\renewcommand{\arraystretch}{1}
\caption{List of atomic meta actions defined for longitudinal and lateral directions. These meta actions represent instantaneous kinematic changes in low-level trajectories at the frame level, in contrast to high-level driving decisions that are composed of multiple atomic actions over a video segment.}
\vspace{-0.2cm}
\begin{tabular}{@{}p{3.5cm}p{3.5cm}p{3.5cm}p{3.5cm}@{}}
\toprule
\multicolumn{2}{c}{\textbf{Longitudinal}} & \multicolumn{2}{c}{\textbf{Lateral}} \\
\midrule
Gentle accelerate & Strong accelerate & Steer left & Steer right \\
Gentle decelerate & Strong decelerate & Sharp steer left & Sharp steer right \\
Maintain speed & Stop & Reverse left & Reverse right \\
Reverse & -- & Go straight & -- \\
\bottomrule
\end{tabular}
\renewcommand{\arraystretch}{1}
\label{tab:meta_actions}
\end{table}

\label{subsec::auto_labeling}
\subsubsection{Auto-Labeling} 

\mypara{Keyframe Selection for Auto-Labeling.} To efficiently scale up training data and enhance model generalization, we develop an auto-labeling pipeline for \datashortname annotation. To identify keyframes for auto-labeling, we first define a set of low-level meta actions and implement corresponding rule-based detectors to infer these meta actions at the frame level. Then, we treat the frame at which a meta action transition occurs as a decision-making moment, allowing us to determine the keyframe automatically and efficiently across large scale data. 

\mypara{Meta Actions.} The complete list of these meta actions is provided in \cref{tab:meta_actions}. These low-level meta actions are \textit{atomic}, representing instantaneous kinematic changes in the ego vehicle's trajectory, and are therefore distinct from high-level driving decisions. A single high-level driving decision within a video segment typically consists of a sequence of such atomic meta actions across both longitudinal and lateral directions. For example, a left lane-change decision may comprise a sequence of \textit{steer left}, followed by a brief \textit{steer right} to stabilize the vehicle heading, and then \textit{go straight}, often accompanied by a \textit{gentle accelerate} and \textit{maintain speed}. For each 8-second data sample, we annotate at most one longitudinal and one lateral high-level driving decision, while atomic meta actions are automatically labeled at 10Hz.

\mypara{Labeling Procedure.} Next, we employ state-of-the-art VLMs such as GPT-5~\citep{gpt5} to perform offline auto-labeling through a multi-step reasoning process. This approach distills world knowledge from large models into structured \datashortname annotations, while balancing efficiency and cost. Similar to the human labeling pipeline, VLMs generate structured reasoning traces consisting of the identified driving decision, critical components, and a concise reasoning trace that links the driving decision to its causal factors. To support the reasoning process, the auto-labeling pipeline provides the model with both raw video and auxiliary signals, including the ego vehicle’s trajectory, dynamic states, and meta actions. The video is sampled at 2 Hz to balance information density with the allowed input token budget within the auto-labeling model's context window.

To mitigate causal confusion, VLMs are prompted to use the 2-second historical video when identifying critical components. The subsequent 6-second future video, along with the ego’s trajectories and meta actions, is then used to resolve multi-modality and determine the corresponding driving decision. During this process, the model ranks the importance of the identified causal factors and retains only those that directly influence the driving decision in the final reasoning trace. 

\subsubsection{Evaluation}

Assessing open-ended text, especially reasoning traces, remains an open challenge in the AV research community, and evaluating causal-effect relationships in \datashortname introduces an additional layer of complexity. Prior datasets have typically relied on one of the following approaches:
\begin{enumerate}[label=(\arabic*)]
\item \textit{Human evaluation} on a small subset of samples. While effective when labelers are properly guided, this approach is not scalable for large-scale evaluation or rapid iteration of labeling pipelines.
\item \textit{Heuristics-based metrics}, such as BLEU~\citep{papineni2002bleu}, METEOR~\citep{banerjee2005meteor} and CIDEr~\citep{vedantam2015cider}. These metrics focus on capturing only shallow text similarity and fail to reflect underlying causal reasoning, making them inadequate for evaluating our \datashortname dataset.
\item \textit{LLM-based auto-evaluation}, which leverages LLMs' capacity to reason about causal relationships and scales effectively to large evaluation sets. However, LLMs are subject to hallucinations, particularly when assessing complex multi-step cause–and–effect chains.
\end{enumerate}
Due to these challenges, prior works often lack a reliable method for reasoning dataset evaluation.

\mypara{\datashortname Evaluation Procedure.} To address these challenges, we adopt a hybrid evaluation strategy that combines human verification with LLM-based auto-evaluation. Specifically, we use GPT-5~\citep{gpt5} as an LLM evaluator and construct a curated evaluation set of 2K samples spanning representative scenarios listed in \cref{tab:odd}. To mitigate hallucination during LLM evaluation, we avoid using free-form text and grading results directly. Instead, we decompose the evaluation process into three structured subtasks covering driving decisions, presence of causal factors, and validity of the cause-and-effect relationship. By reformulating these aspects as a set of True/False questions, the evaluation process becomes more interpretable and better aligned with human judgment. To validate reliability, we compare LLM-based auto-evaluation against human evaluation on the same version of the auto-labeled dataset, and observe a 92\% alignment rate, confirming the robustness of our LLM-based auto-evaluation. With this evaluation method, we find that the proposed structured \datashortname reasoning traces improve the causal relationship score by 132.8\% relative to free-form reasoning traces, which do not enforce explicit driving decisions and critical components.

\mypara{Effectiveness of Imperfect Auto-Labels.} It is important to note that attaining a perfect (100\%) score in causal-effect evaluation, were it even possible, is not a necessary condition for the usefulness of auto-labeled data. Given the inherent ambiguity of causal reasoning in complex driving scenarios, as well as noise in both human-labeled ground truth and evaluation metrics, it is unclear whether 100\% agreement is a reasonable or well-defined target. Instead, the primary value of \datashortnamenosp’s auto-labels lies in enabling large-scale SFT, which improves \reasoningvlashort's generalization across diverse driving scenarios. Empirically, as will be shown in \cref{sec::experiment}, models trained on auto-labeled \datashortname traces already achieve significant improvements over baselines without reasoning supervision. Moreover, as will be described in \cref{sec::training}, our training pipeline incorporates subsequent RL-based post-training steps which further strengthen reasoning capability and causal consistency. In parallel, as our human annotation effort scales, we plan to introduce additional rounds of SFT using human-labeled \datashortname reasoning traces, progressively improving causal grounding and interpretability.

\section{Training Strategy}
\label{sec::training}

\begin{figure}[t]
    \centering
    \includegraphics[width=\linewidth]{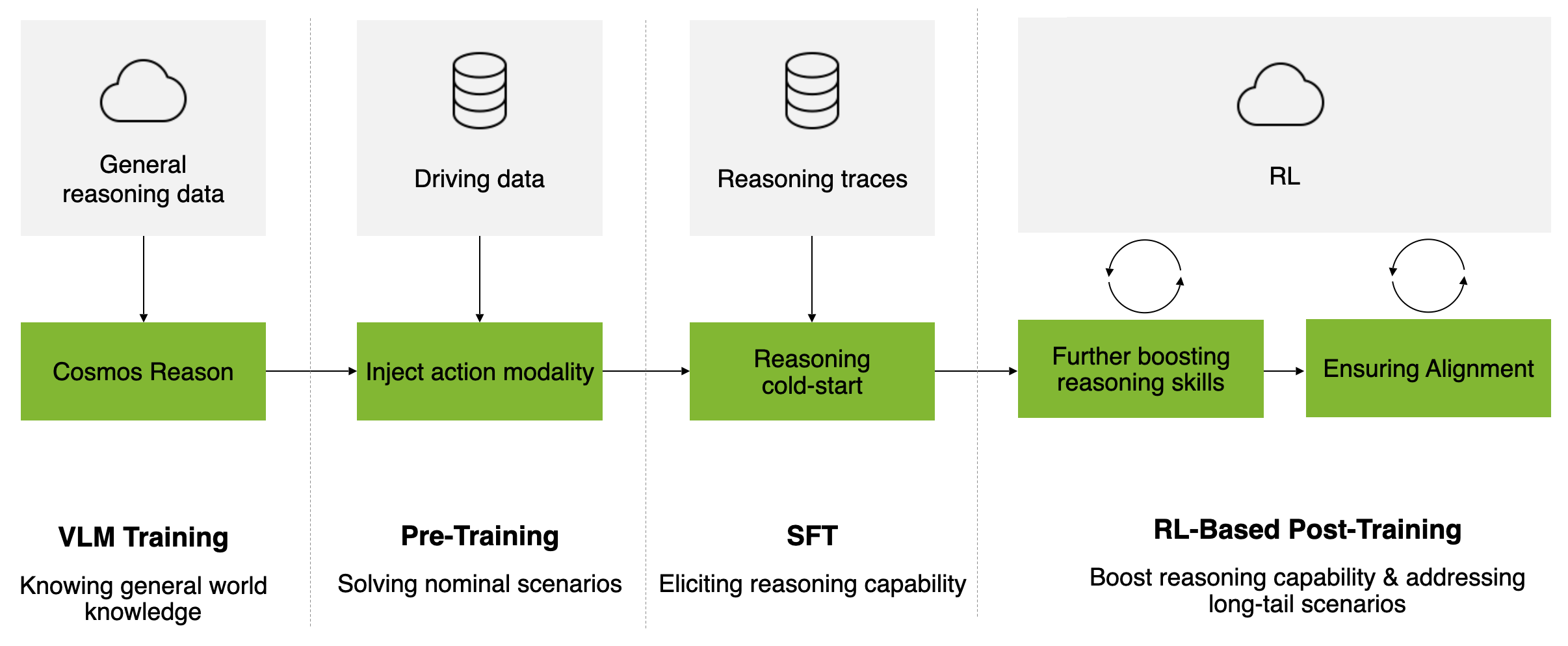}
    \vspace{-0.6cm}
    \caption{Overview of the \reasoningvla model training pipeline, consisting of three key stages: (1) Action Modality Injection (\cref{sec:traj_inject}), (2) Eliciting Reasoning (\cref{sec:elicit_reasoning}), and (3) RL-Based Post-Training (\cref{sec:rl_alignment}).
    }
    \label{fig:train}
\end{figure}

Building upon the Cosmos-Reason VLM backbone introduced in \cref{sec::model}, which provides foundational physical reasoning capabilities through domain-specific SFT, we adopt a three-stage training strategy to transform the VLM into a reasoning-capable autonomous driving policy.
As illustrated in \cref{fig:train}, each stage progressively enhances distinct capabilities essential for robust and interpretable driving.
In \cref{sec:traj_inject}, we inject the action modality into the VLM by training with discrete trajectory tokens and adding an action-expert based on flow matching, enabling the model to predict vehicle control outputs.
In \cref{sec:elicit_reasoning}, we improve the model's reasoning capability through SFT on the \datashortname dataset (\cref{sec::data_labeling}), teaching the model to generate causally grounded explanations for better driving decisions. Finally, in \cref{sec:rl_alignment}, we employ RL with large reasoning model feedback to refine reasoning quality, align reasoning traces with executed actions, and optimize trajectory quality, producing interpretable and safe driving behavior. 

\subsection{Action Modality Injection}
\label{sec:traj_inject}

During training, we inject the action modality to the VLM through discrete tokens (\cref{sec::action_decoding}) and train the VLM via cross-entropy loss over the training token sequence defined in \cref{eq:training_seq}.  
Following the control-based representation in \cref{eq:dyn_trajectory}, each trajectory consists of 64 waypoints with 2 quantized values per waypoint (acceleration $a^i$ and curvature $\kappa^i$), resulting in 128 discrete tokens per trajectory. 
These are encoded with a set of special tokens dedicated to action representation. 
However, we do not use discrete trajectory tokens for inference, as detailed below.

\textbf{Motivation for Dual Representation.} The use of discrete tokens during training alongside a continuous flow-matching decoder at inference provides several key advantages. 
First, discrete tokenization enables \textit{unified} autoregressive training in which reasoning and trajectories share a common token space, allowing the VLM to tightly couple causal explanations with vehicle behaviors through standard next-token prediction. 
Second, discrete representations facilitate RL optimization by allowing direct gradient flow during post-training (\cref{sec:rl_alignment}), allowing policy gradient methods such as GRPO~\citep{shao2024deepseekmath} to jointly refine reasoning quality and reasoning-action consistency.  Third, the discrete representation provides strong supervision for learning vehicle dynamics, while the flow-matching expert ensures physically feasible and multi-modal outputs. Finally, flow-matching decoding offers computational efficiency, generating continuous trajectories substantially faster than autoregressively sampling 128 discrete tokens, enabling real-time inference.

Similar to $\pi_{0.5}$-KI~\citep{driess2025knowledge}, we adopt a separate action-expert to decode actions via flow matching~\citep{pmlr-v162-janner22a,lipman2023flow,ctg,jiang2023motiondiffuser}.
The action-expert follows the same Transformer architecture as the VLM, using the same number of attention heads and attention dimensions, but with a smaller hidden embedding size and MLP dimension for efficiency.
At each diffusion timestep $t$ in the diffusion schedule, the action-expert takes as input both the KV-cache from the sequence $[\observation_\text{image}, \observation_\text{egomotion}, \reasoning]$ in the VLM and the embedded representation of the noisy control $\dyntraj_t$ (with the diffusion time $t$ also embedded and added to the feature).
The expert then predicts the vector field $\mathbf{v}_\Theta(\dyntraj_t, \bm{o}, \reasoning)$ by projecting the final layer feature through an MLP head, where $\Theta$ denotes the learnable parameters.
We train the action-expert using a vanilla conditional flow matching loss~\citep{lipman2023flow},
\begin{equation}
    L_\text{cfm}(\Theta) = \mathbb{E}_{t \in p_\text{schedule}, (\observation,\reasoning) \in \mathcal{D}_\text{data}}\|\mathbf{v}_\Theta(\dyntraj_t, \bm{o}, \reasoning) - \mathbf{u}(\dyntraj_t|\dyntraj) \|.
\end{equation}
In practice, we adopt the Gaussian conditional optimal transport (OT) path and sample $\dyntraj_t = t\dyntraj + (1-t)\bm{\epsilon}$ with $\bm{\epsilon} \sim \mathcal{N}(\bm{0}, \bm{I})$, where the target vector field admits a closed-form expression:
\begin{equation}
\mathbf{u}(\dyntraj_t|\dyntraj) = \dyntraj - \bm{\epsilon}.
\end{equation}
During inference, starting with $\dyntraj_0 \in \mathcal{N}(\bm{0}, \bm{I})$, we perform denoising through Euler integration:
\begin{equation}
    \dyntraj_{t+\delta_t} = \dyntraj_t + \delta_t \, \mathbf{v}_\Theta(\dyntraj_t, \observation, \reasoning).
\end{equation}
By default, we use $\delta_t = 0.1$ during inference and set $p_\text{schedule}$ to a shifted beta distribution during training, as suggested by \citet{intelligence2504pi0}. During training, we apply a stop-gradient to the KV-cache produced by the VLM to prevent gradients from the expert back-propagating into the VLM weights.

\subsection{Eliciting Reasoning}
\label{sec:elicit_reasoning}

Having established a VLA with action generation capabilities in \cref{sec:traj_inject}, the next challenge is to enable the model to perform structured and causally grounded reasoning that explains \textit{why} specific driving decisions are made. This capability is critical for handling complex, safety-critical scenarios where pure pattern matching from imitation learning may fail~\citep{wei2022chain}. To achieve this, we leverage the structured \datashortname dataset introduced in \cref{sec::data_labeling}, which provides decision-grounded and causally linked reasoning traces paired with expert trajectories. We perform SFT on the \datashortname dataset to teach the model to generate reasoning traces through imitation, where each reasoning trace is anchored to explicit driving decisions (\cref{tab:driving_decisions}) and grounded in critical scene components (\cref{tab:critical_components}). While SFT enables the model to scaffold basic reasoning capabilities, we further refine reasoning quality and enforce reasoning-action consistency through RL in \cref{sec:rl_alignment}.
Formally, each training sample consists of a multi-camera driving scene observation $\observation = [\observation_\text{image}, \observation_\text{egomotion}]$, a structured \datashortname reasoning trace $\reasoning$ that explains the causal factors behind the ego vehicle's decision, and the corresponding ground-truth control-based trajectory representation $\dyntraj$ defined in \cref{eq:dyn_trajectory}. Following the sequence formulation in \cref{eq:training_seq}, the SFT objective maximizes the conditional log-likelihood of the reasoning–action sequence:
\begin{equation}
    \mathcal{L}_{\text{SFT}}(\theta) = -\mathbb{E}_{(\observation,\reasoning, \dyntraj)\sim\mathcal{D}_{\text{CoC}}} \left[\log \pi_{\theta} (\reasoning,\,\dyntraj \mid \observation)\right],\label{eq:sft_coc_loss}
\end{equation}
where $\pi_{\theta}$ denotes the VLA policy parameterized by $\theta$, encompassing the vision encoder, language backbone, and corresponding embedding adapters.
In practice, we apply the cross-entropy loss over both the reasoning tokens and the discrete trajectory tokens (128 tokens per trajectory as described in \cref{sec:traj_inject}), enabling the model to learn the joint distribution of language-based reasoning and action prediction in a unified autoregressive framework.

\textbf{Why SFT Alone is Insufficient.}
This imitation learning stage allows the model to internalize human-like reasoning patterns:  learning not only \emph{what} action to take, but also \emph{why} such actions are appropriate given specific visual and contextual cues. As shown in \cref{fig:cot_open_loop_results}, SFT on \datashortname data already yields measurable improvements in trajectory prediction accuracy compared to models trained without explicit reasoning supervision. However, while SFT enables the VLA model to scaffold reasoning traces, it remains inherently limited by several factors: 
\begin{enumerate}[label=(\arabic*), leftmargin=*, itemsep=2pt]
    \item \textit{Data bias and annotation noise}: Auto-labeled data may contain imperfect causal relationships (\cref{subsec::auto_labeling}), causing the model to overfit to annotation artifacts rather than learning robust causal reasoning.
    \item \textit{Limited generalization}: The model may memorize common reasoning patterns without developing deeper causal understanding, failing to generalize to novel scenarios.
    \item \textit{Weak visual grounding}: Next-token prediction does not enforce visual consistency; the model may hallucinate causal factors not present in the scene (\cref{fig:reasoning_reward_demo}).
    \item \textit{Reasoning–action inconsistency}: Joint optimization does not explicitly enforce alignment between stated reasoning and predicted trajectories, potentially leading to contradictory explanations (\cref{fig:reasoning_consistency_reward_demo}).
\end{enumerate}

In the next section (\cref{sec:rl_alignment}), we illustrate our approach to mitigate these limitations via RL-based post-training with large reasoning model feedback and explicit reasoning-action consistency rewards.

\subsection{RL-based Post-Training}
\label{sec:rl_alignment}

\begin{figure}[t]
    \centering
    \includegraphics[width=\linewidth]{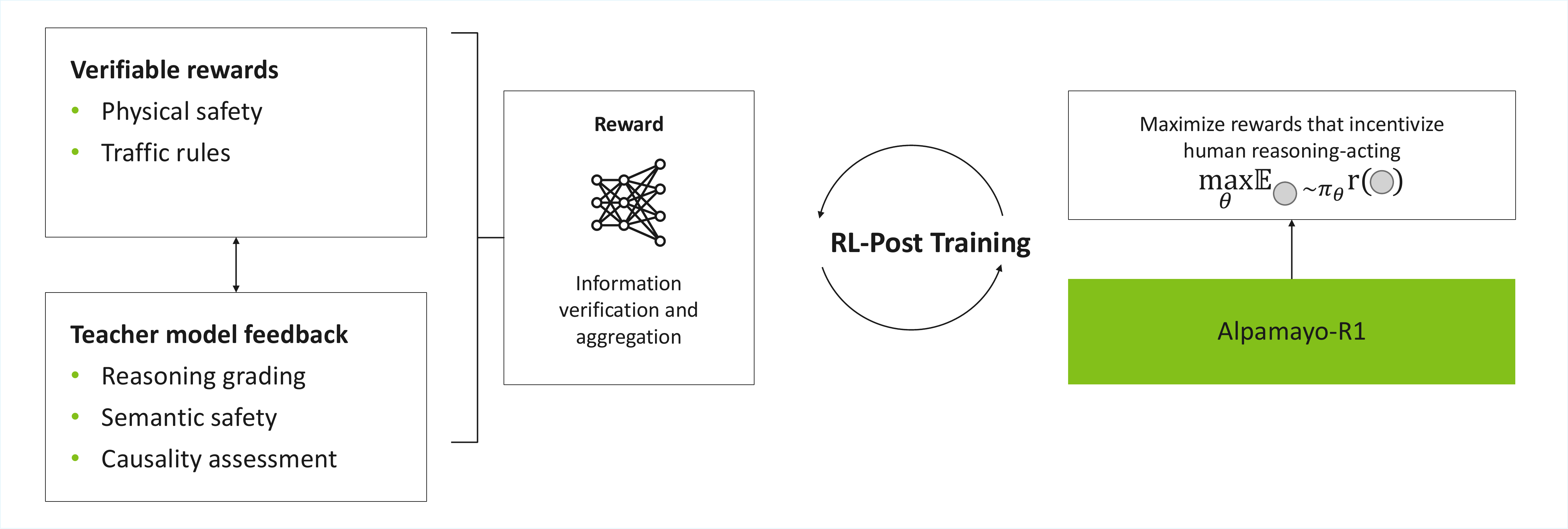}
    \vspace{-0.7cm}
    \caption{Overview of our RL-based post-training framework. We optimize three reward components: reasoning quality (via large reasoning model feedback), reasoning-action consistency, and trajectory quality, to align the model's generated reasoning traces with its predicted actions.}
    \label{fig:rl_training}
\end{figure}

To address the limitations of SFT outlined in \cref{sec:elicit_reasoning}, we introduce an RL-based post-training framework shown in \cref{fig:rl_training} that optimizes three complementary reward signals: reasoning quality (via large reasoning model feedback), reasoning-action consistency, and trajectory quality. Unlike SFT, which optimizes the likelihood of expert demonstrations under \emph{teacher forcing} without feedback on the test-time inference errors, RL provides explicit inference feedback on the model’s own rollouts, aligning the optimization objective with how the system is actually deployed. This approach directly tackles the shortcomings of SFT by providing targeted feedback that evaluates both the causal correctness of reasoning and its alignment with executed actions, and yields disproportionately larger gains in robustness and generalization for the same compute budget.

\subsubsection{Post-Training Algorithm}

Large-scale foundation model post-training has emerged as a central strategy to enhance the reasoning capabilities and generation quality of large-scale foundation models \citep{christiano2017deep, deepseekai2025deepseekr1}.
Recently, these techniques have been extended to the embodied AI domain, encouraging VLA models to generate actions that better reflect human intent across diverse embodiments, including autonomous driving \citep{tian2025direct} and generalist robotic agents \citep{tian2024maximizing, zhang2025rewind}.
In our \textit{reasoning} VLA context, the alignment stage extends beyond improving motion generation; it explicitly enhances reasoning quality grounded in embodied settings and enforces reasoning–action consistency, both of which are key properties for achieving interpretable and trustworthy autonomy.

We adopt GRPO~\citep{shao2024deepseekmath} as our alignment algorithm. GRPO extends standard policy gradient methods by optimizing relative advantages within a group of sampled model rollouts rather than relying on absolute reward signals. 
Specifically, given a group of model rollouts $\{\tau_i\}_{i=1}^K$ sampled from the current model $\pi_\theta$, each with an associated scalar reward $r_i$, the objective of GRPO is defined as:
\begin{equation}
\mathcal{L}_{\text{GRPO}}(\theta)
= - \mathbb{E}{\tau_i \sim \pi_\theta}
\left[
\frac{\exp(\beta A_i)}{\sum_j \exp(\beta A_j)}
\left(\log \pi_\theta(\tau_i) - \lambda_{\mathrm{KL}}\mathrm{KL}[\pi_\theta(\tau_i)\|\pi_{\text{ref}}(\tau_i)]\right)
\right],
\quad A_i = r_i - \bar{r}.  
\end{equation}

Here, $A_i$ denotes the relative advantage of each trajectory within the group, $\bar{r}$ is the group-average reward, and $\beta$ controls the sharpness of the weighting distribution.
The KL regularization term with coefficient $\lambda_{\mathrm{KL}}$ penalizes deviations from the reference policy $\pi_{\text{ref}}$ (typically the SFT model), preventing over-optimization on noisy or biased reward signals and preserving linguistic and behavioral priors learned during pre-training.

\subsubsection{Reward Model}
\label{sec:rl_alignment_reward}

Our reward model integrates three complementary signals that together evaluate both what the model reasons and how it acts. Specifically, the total reward $r$ for each rollout is composed of three components: reasoning quality reward, reasoning-action consistency, and low-level trajectory quality.

\textbf{Grading Reasoning with Large Reasoning Models.} 
To mitigate the issue where reasoning traces can exhibit hallucinations that produce plausible yet unsafe or causally inconsistent plans, we employ large reasoning models (LRMs) as automatic evaluators to provide scalable, high-quality feedback on reasoning quality. Inspired by recent advances in LLM alignment, where expert models serve as judges to provide scalable feedback~\citep{bai2022constitutional,leerlaif}, we leverage state-of-the-art LRMs (e.g., DeepSeek-R1~\citep{deepseekai2025deepseekr1}, Cosmos-Reason~\citep{nvidia2025cosmosreason1physicalcommonsense}) as \emph{reasoning critics} to evaluate the quality of reasoning traces generated by the VLA.
We choose an LRM as the critic because, although such models may struggle to generate driving-specific reasoning due to limited embodiment priors, they exhibit strong verification and evaluation capabilities. In other words, even when generation in this domain is imperfect, their ability to assess logical soundness, causal alignment, and contextual consistency remains highly reliable (also known as the generation–verification gap \citep{song2024mind}).
The resulting reward signal provides a continuous measure of reasoning quality, enabling RL to iteratively refine the model's ability to generate grounded and logically consistent reasoning.

\textbf{Reasoning Critic Design.}
For each training sample, the LRM critic takes as input the multi-camera visual observation $\observation_\text{image}$ at the last frame of the 2-second history window, the ground-truth \datashortname reasoning trace $\reasoning_{\text{GT}}$ from the dataset, and the model-generated reasoning trace $\reasoning_{\text{pred}}$ produced by the current policy $\pi_\theta$. 
The critic evaluates how well $\reasoning_{\text{pred}}$ aligns with $\reasoning_{\text{GT}}$ along two dimensions: \textit{behavior consistency}, whether the predicted reasoning describes a driving decision consistent with ground truth; and \textit{causal reasoning quality}, whether it correctly identifies causal factors observable in the scene's history according to \datashortname principles (\cref{sec:structured_coc}).
The critic grades the predicted reasoning according to a structured rubric focused on behavior consistency and causal reasoning consistency:
\begin{promptbox}[prompt:reasoning-eval]{LLM Reasoning Grading Rubric}
\small
You are an expert evaluator for autonomous driving reasoning traces. The reasoning trace describes what the ego vehicle should be doing and the reasons and factors that lead to the behavior. Your task is to score how well a predicted reasoning trace (\texttt{PRED}) aligns with the ground truth (\texttt{GT}) in terms of behavior consistency and causal reasoning.

\textbf{Scoring rubric (0–5):}
\begin{itemize}
    \item[5] Behavior \& causal reasoning fully consistent.
    \item[4] Behavior correct; causal reasoning mostly consistent.
    \item[3] Behavior roughly correct, but incomplete or slightly incorrect reasoning.
    \item[2] Behavior partially incorrect or reasoning largely inconsistent.
    \item[1] Behavior is wrong or contradicts GT.
    \item[0] Completely unrelated or opposite.
\end{itemize}
\end{promptbox}

The resulting scalar score $r_{\text{reason}}$ is used as the reasoning reward. This signal encourages the model to generate reasoning traces that not only describe correct driving behaviors but also maintain causal fidelity, accurately explaining why an action is taken based on visual context and traffic cues.

\textbf{\datashortnamenosp-Action Consistency}. To ensure that the model’s action generation faithfully follows its reasoning, we introduce a \datashortnamenosp–action consistency reward that measures behavioral alignment between the generated reasoning trace and the corresponding predicted ego trajectory.
Specifically, for each reasoning–action rollout, we convert the predicted motion trajectory into a sequence of meta-actions (interpretable motion primitives) described in \cref{subsec::auto_labeling}.
These meta-actions encode the ego vehicle’s control behavior along both the longitudinal (acceleration/braking) and lateral (steering) directions. We then parse the generated reasoning trace to infer the ego’s intended behavior and compare it against the meta-actions derived from the predicted trajectory using rule-based matching. If the described behavior in the reasoning trace and the meta-action are consistent across both axes, we assign $r_{\text{consistency}} = 1$; otherwise, $r_{\text{consistency}} = 0$. In cases where the reasoning cannot be parsed into a valid driving decision (i.e., the intent is not recognized within the closed decision set used for auto-labeling), we conservatively assign $r_{\text{consistency}} = 0$.
Although based on simple rule-based logic, this binary reward plays a crucial role in improving the trustworthiness of the model's reasoning–action coupling.
By explicitly penalizing inconsistencies and rewarding only correct matches, it encourages the model to generate reasoning that not only sounds plausible but also translates into coherent, physically consistent behavior.

\textbf{Low-Level Trajectory Quality.} To ensure that the generated motion trajectories remain physically feasible, comfortable, and safe to execute, we include a low-level trajectory quality reward that evaluates the model’s motion outputs in continuous space.
This component complements the above reasoning- and consistency-level rewards by directly regularizing the trajectory’s physical properties.
The reward combines three terms:
\begin{equation}
r_{\text{traj}} = \lambda_{\text{L2}} \|x_{\text{pred}} - x_{\text{expert}}\|_2^2 + \lambda_{\text{coll}} \mathbb{I}[\text{collision}(x_{\text{pred}})] + \lambda_{\text{jerk}}J(x_{\text{pred}}), 
\end{equation}
where $x_{\text{pred}}$ and $x_{\text{expert}}$ denote the predicted and expert trajectories, respectively;
$\mathbb{I}[\text{collision}(x_{\text{pred}})]$ is a binary indicator that denotes whether the predicted motion leads to a collision with surrounding obstacles;
and $J(x_{\text{pred}})$ measures the magnitude of the jerk to penalize abrupt or uncomfortable motion.
The L2 imitation term encourages proximity to expert demonstrations, promoting stable learning and smooth driving profiles. The collision penalty ensures safety, while the jerk regularization improves comfort and control smoothness.
Together, these terms anchor the learning of the model to human-like, safe, and comfortable motion, reinforcing the physical plausibility of the trajectories generated during the alignment process.

\begin{figure}[t]
    \centering
    \includegraphics[width=0.7\linewidth]{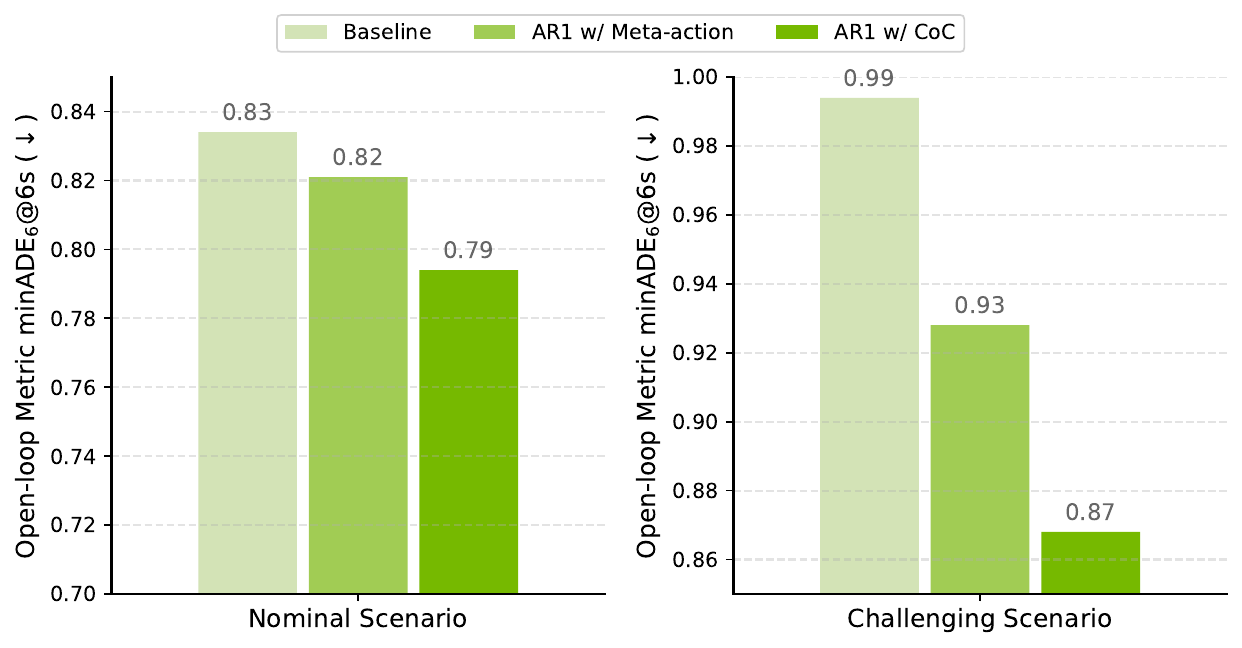}
    \vspace{-0.2cm}
    \caption{Compared to models that only output trajectories or only output meta-actions and trajectories, \reasoningvla achieves improvements in both nominal and challenging scenarios.}
    \label{fig:highlight_results}
\end{figure}

\subsubsection{Post-Training Data Curation for Cost-Effective Training}
RL–based post-training is computationally expensive due to its iterative nature: each policy update requires multiple model rollouts, reward evaluations, and gradient steps across large batches of reasoning and trajectory samples.
Moreover, unlike the SFT stage where the loss is directly computed from labeled data, our post-training procedure involves on-policy sampling and LRM-based reward function calls, which amplify both compute and data costs.
Consequently, scaling RL to the full pre-training data would be prohibitive in both training time and compute resources.
To address this, we curate a high-information-gain dataset for RL post-training.
The key idea is to prioritize samples where the model’s implicit reward signal (encoded in its logits) disagrees with the explicit reward model.

Specifically, for each sample rollout from the model (denoted as $\tau_i$), we compute the model’s predicted probability distribution derived from its logits, and the corresponding probability distribution implied by the rewards, which we obtain by transforming the reward into a Boltzmann distribution
$p_{\text{reward}}(\tau_i) = \frac{\exp(\beta \, r_i)}{\sum_j \exp(\beta \, r_j)}$.
A large divergence between these two distributions indicates that the model’s internal preference (its implicit reward) conflicts with the externally defined reward signal. Such disagreement reveals samples where the model’s learned reward is inaccurate, making them particularly valuable for alignment. We therefore prioritize these high-disagreement samples to construct a focused post-training dataset, while mixing in a similar proportion of randomly sampled data to preserve distributional diversity and stabilize training. By focusing RL updates on this hybrid set, we achieve both high alignment efficiency and robust learning dynamics compared to uniformly sampled data.

\subsubsection{Post-Training Infrastructure}

To conduct our RL experiments, we develop a customized version of the Cosmos-RL framework \citep{cosmosrl2025} that is specifically designed for AV reasoning tasks. This system provides a scalable, modular infrastructure for large-scale multimodal RL and fits directly with other parts of the \reasoningvla system. It supports distributed data loading, mixed-parallelism training, vLLM-based rollout generation~\citep{vllm}, and reward computation across multiple GPU nodes, enabling efficient, high-throughput policy optimization.

\section{Experiments}
\label{sec::experiment}

We conduct comprehensive evaluations of \reasoningvla across multiple dimensions to assess its reasoning capabilities, trajectory prediction accuracy, and closed-loop driving performance. We first highlight in \cref{fig:highlight_results} that the proposed \reasoningvla significantly outperforms the trajectory-only baseline, particularly in challenging scenarios that intuitively require complex reasoning to make better driving decisions.

In the following sections, we first present the evaluation protocol in \cref{subsec::protocol}. Next, we illustrate how our reasoning-capable model contributes to an improved driving policy in \cref{subsec::reasoning_results}. In \cref{subsec::alignment_results} we further demonstrate the improvements in behavioral alignment achieved through RL. From \cref{subsec::abl_backbone} to \cref{sec::exp::visenc} we conduct a comprehensive ablation study on the backbone model, the trajectory expert model, and the vision encoder to gain deeper insight into the effectiveness of our proposed methodology. Finally, we present an on-vehicle demonstration showcasing the real-world performance of our model.

\subsection{Evaluation Protocol}
\label{subsec::protocol}

\definecolor{myblue}{HTML}{D0E3FA}
\definecolor{mygreen}{HTML}{c5e096}
\definecolor{mygray}{HTML}{e3e3e3}
\begin{table}[t]
    \centering
    \caption{Open-loop evaluation of models on the CoC dataset. The base model is pre-trained with $\mathcal{D}_{\text{overall}}$ and all other models are finetuned on the CoC dataset,  then evaluated on held-out CoC test data. Numbers with green background are the best under each setting.
    }
    \begin{tabular}{clccccc}
        \toprule
        \textbf{ID} & \textbf{Model Name} & \textbf{Route} & \textbf{Parameters} & \textbf{minADE$_6$@3s}$\downarrow$  & \textbf{minADE$_6$@6.4s}$\downarrow$  \\
        \midrule
        1 & Base model (action modality)   & $\times$ & 0.5B         & 0.284    & 0.996    \\
        2 & + Ft. w/ Traj.                  & $\times$   & 0.5B      & 0.282    & 0.971    \\
        3 & + Ft. w/ Meta-action \& Traj.   & $\times$   & 0.5B      & 0.291    & 0.988    \\
        \rowcolor{mygreen}
        4 & + Ft. w/ CoC \& Traj. (\reasoningvlashort)     & $\times$   & 0.5B      & 0.279    & 0.955    \\
        \midrule
        5 & Base model (action modality)    & $\times$   & 3B        & 0.291   & 0.977  \\
        6 & + Ft. w/ Traj.                  & $\times$   & 3B        & 0.293   & 0.976   \\
        7 & + Ft. w/ Meta-action \& Traj.  & $\times$   & 3B        & 0.280   & 0.927   \\
        \rowcolor{mygreen}
        8 & + Ft. w/ CoC \& Traj. (\reasoningvlashort)    & $\times$   & 3B        & 0.275   & 0.908    \\
        \midrule
        9 & Base model (action modality)     & $\checkmark$ & 0.5B                & 0.264    &   0.848  \\
        10 & + Ft. w/ Traj.                  & $\checkmark$   & 0.5B               & 0.262   & 0.834    \\
        11 & + Ft. w/ Meta-action \& Traj.  & $\checkmark$   & 0.5B                & 0.264    & 0.821    \\
        \rowcolor{mygreen}
        12 & + Ft. w/ CoC \& Traj. (\reasoningvlashort)    & $\checkmark$   & 0.5B    & 0.254  & 0.794     \\
        \bottomrule
    \end{tabular}
\label{tab:cot_open_loop_results}
\end{table}

\begin{table}[t]
    \centering
    \caption{Open-loop evaluation of models on the challenging dataset. All models are finetuned on the CoC dataset and evaluated on the challenging dataset.
    }
    \begin{tabular}{clccccc}
        \toprule
        \textbf{ID} & \textbf{Model Name} & \textbf{Route} & \textbf{Parameters} & \textbf{minADE$_6$@3s}$\downarrow$  & \textbf{minADE$_6$@6.4s}$\downarrow$  \\
        \midrule
        1 & Ft. w/ Traj.                  & $\checkmark$   & 0.5B      & 0.315    & 0.994    \\
        2 & Ft. w/ Meta-action \& Traj.   & $\checkmark$   & 0.5B      & 0.301   & 0.928   \\
        \rowcolor{mygreen}
        3 & Ft. w/ CoC \& Traj. (\reasoningvlashort)  & $\checkmark$   & 0.5B      & 0.290    & 0.868    \\
        \bottomrule
    \end{tabular}
\label{tab:cot_open_loop_results_challenging}
\end{table}

\begin{figure}[htbp]
    \centering
    \includegraphics[width=1.0\linewidth]{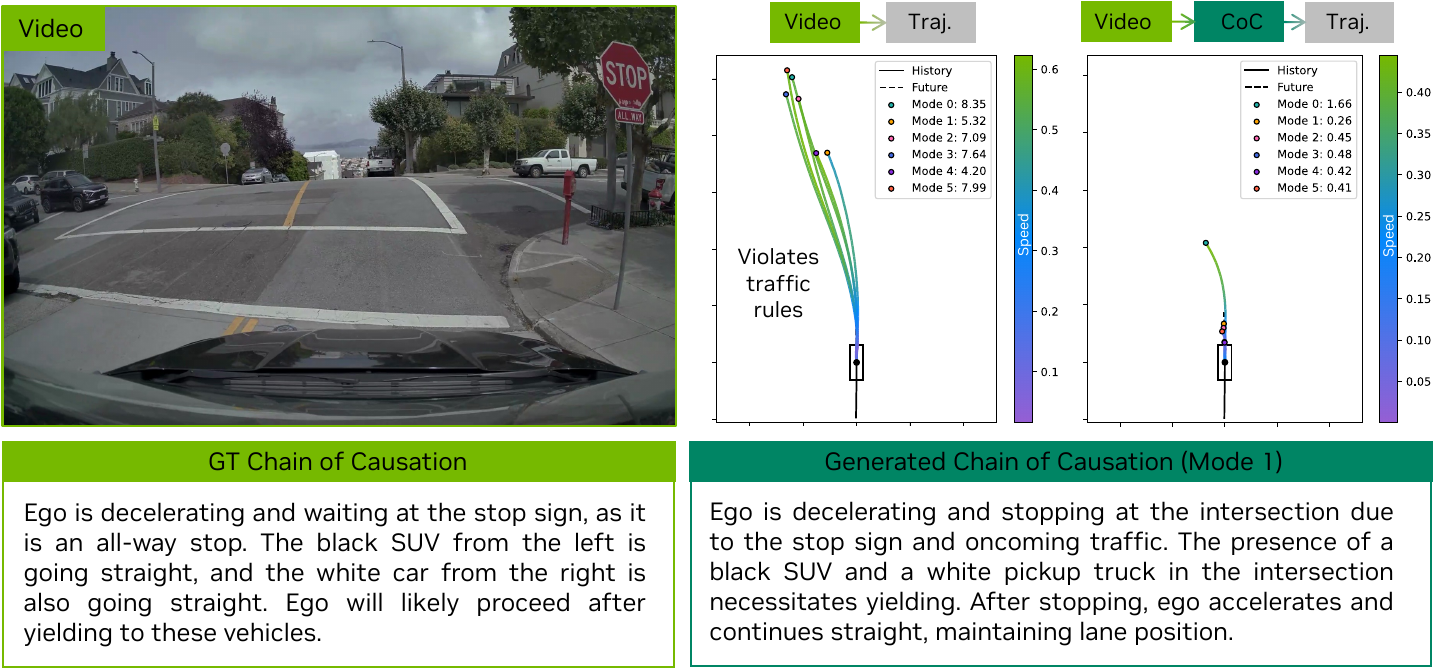}
    \vspace{-5mm}
    \caption{Policy improvements via eliciting reasoning: \reasoningvla generates a correct reasoning trace at an all-way stop sign intersection and yields to other vehicles that enter the intersection earlier than ego.}
    \label{fig:cot_open_loop_results}
\end{figure}

Our evaluation strategy consists of four complementary components:
\begin{enumerate}[label=(\arabic*)]
\item \textit{open-loop trajectory prediction} on both nominal and long-tail driving scenarios to measure planning accuracy;
\item \textit{closed-loop simulation} using AlpaSim~\citep{alpasim2025} to assess safety and robustness when the model controls the vehicle in realistic scenarios;
\item \textit{ablation studies} examining the impact of key architectural choices, including vision-language model scaling, vision encoding strategies, reasoning integration, and action decoding strategies;
\item \textit{on-vehicle road tests} to validate real-world deployment of the model in autonomous driving scenarios.
\end{enumerate}

\textbf{Dataset.} We train and evaluate models on internal driving data collected across diverse geographic regions in the US and EU, with all evaluation data strictly geo-fenced and held out from training regions to prevent information leakage. 
Our evaluation encompasses both nominal driving scenarios in dataset $\mathcal{D}_{\text{overall}}$ and challenging long-tail cases in $\mathcal{D}_{\text{hard}}$ to thoroughly test the model's ability to handle rare, safety-critical events. In detail, the full training and evaluation dataset comprises 80,000 hours of driving data collected from multiple ego-vehicles operating in more than 2,500 cities in 25 countries. It encompasses diverse driving scenarios, including highway and urban environments, under various weather conditions, times of day, and traffic densities. The raw sensory inputs consist of video recordings from a surround-view seven-camera setup, accompanied by precise camera calibration parameters and ego-motion data.  In this work, we focus on using two front-facing cameras as input: a front wide-angle camera with 120$^{\circ}$ field of view and a front telephoto camera with 30$^{\circ}$ field of view, providing complementary perspectives for both near-field and far-field scene understanding. 

In addition to the general driving dataset $\mathcal{D}_{\text{overall}}$, we construct the \datashortname dataset (\cref{sec::data_labeling}) consisting of 700K video segments with structured CoC. This dataset is used for fine-tuning models to elicit reasoning capabilities (\cref{subsec::reasoning_results}) and for RL-based post-training alignment (\cref{subsec::alignment_results}).

\textbf{Open-Loop Evaluation.}
For open-loop trajectory prediction, we evaluate models over a prediction horizon of 6.4 seconds, corresponding to the ego-vehicle's planned waypoints. We use minADE and ADE as the evaluation metric. minADE is computed over 6 samples ($\text{minADE}_6$) and is defined as the minimum distance between the ground-truth future trajectory and the best-matching trajectory among 6 predictions generated by the model. ADE (Average Displacement Error) is the average distance between the predicted trajectory and the ground-truth trajectory across all future timesteps.

\textbf{Closed-Loop Evaluation.}
It is well established that strong open-loop results do not necessarily translate into reliable closed-loop driving performance~\citep{dauner2023parting}. To address this gap, we further evaluate our models within \textit{AlpaSim}~\citep{alpasim2025}, an open-source closed-loop end-to-end simulator based on state-of-the-art neural reconstruction technology~\citep{wu20253dgut}. AlpaSim leverages a temporal 3D Gaussian Splatting representation from recorded real-world driving logs and, during closed-loop evaluation, uses it to synthesize novel viewpoints when the ego vehicle deviates from the recorded trajectory. During evaluation, predicted trajectories are tracked by a model predictive controller (MPC), and vehicle dynamics follow a dynamically extended bicycle model. Traffic agents, including vehicles and pedestrians, follow their recorded trajectories. 

We evaluate models in 75 challenging 20-second scenarios, selected for their dense ego–agent and agent–agent interactions. While this may appear as a limited set, these scenarios are specifically curated to represent the most demanding safety-critical situations requiring complex reasoning and interactive decision-making.
We report the following AlpaSim metrics:
\begin{enumerate}[label=(\arabic*)]
\item \textit{close encounter rate (all)}: percentage of scenarios where the ego vehicle experiences a close encounter with any other traffic agent;
\item \textit{close encounter rate (at-fault)}: same as close encounter rate but considering only close encounters  where the ego vehicle is deemed responsible, i.e., excluding rear-end close encounters.
\item \textit{offroad rate}: percentage of scenarios where the ego vehicle drives outside of the drivable area;
\item \textit{AlpaSim score (all)}: average distance driven in km between events, where events correspond to offroad or close encounter occurrences;
\item \textit{AlpaSim score (at-fault)}: same as AlpaSim score but considering only close encounters  where the ego vehicle is deemed responsible, i.e., excluding rear-end close encounters.
\end{enumerate}
The simulation ends after the first close encounter or off-road event. To mitigate rendering artifacts, events in which the ego deviates more than 4\,m from the original recorded trajectory are excluded from all metric computations.

\subsection{Policy Improvements from Reasoning}
\label{subsec::reasoning_results}

One of the key contributions of this work is the use of the proposed CoC data to improve driving policies.
To evaluate the impact of reasoning on driving performance, we start with a base model pre-trained on $\mathcal{D}_{\text{overall}}$ with action modality injection (\cref{sec:traj_inject}), then fine-tune it on the \datashortname dataset with different reasoning modalities: meta-action descriptions and full \datafullnamelower
reasoning traces. During inference, models trained with \datashortname reasoning generate explicit reasoning outputs alongside trajectory predictions, enabling them to better handle challenging scenarios that require multi-step decision making.
We compare three fine-tuning strategies: (1) trajectory prediction only, (2) meta-action and trajectory prediction, and (3) \datafullnamelower reasoning and trajectory prediction (\reasoningvla). All models are evaluated on held-out \datashortname test data in two settings: with and without route information provided to the model. 

\textbf{Open-Loop Improvements.} As shown in \cref{tab:cot_open_loop_results} (nominal scenarios) and \cref{tab:cot_open_loop_results_challenging} (challenging scenarios), incorporating \datashortname reasoning yields substantial improvements in open-loop trajectory prediction in both settings. Without route information, \reasoningvlashort achieves a minADE$_6$ of 0.955m at 6.4s, a 4.1\% improvement over the base model and outperforming both trajectory-only (0.971m) and meta-action (0.988m) baselines. 
With route information, the gains are more pronounced: \reasoningvlashort achieves 0.794m, representing 4.8\% improvement over the trajectory-only baseline (0.834m). Scaling to 3B parameters further improves performance, with \reasoningvlashort-3B achieving 0.908m (without route), demonstrating the benefits of increased model capacity for complex reasoning tasks. In challenging scenarios, the improvements are even larger, with \reasoningvlashort achieving 0.868m, a \textit{12\% improvement} over the trajectory-only baseline (0.994m).

These results demonstrate that explicit reasoning capabilities enable the model to more effectively leverage contextual information such as route guidance and handle complex driving scenarios that require anticipating future interactions. \cref{fig:cot_open_loop_results} illustrates qualitative examples where the CoC-enabled model successfully generates correct reasoning traces and yields to vehicles in challenging scenarios, while baseline models fail to anticipate these interactions.

\begin{figure}[t]
    \centering
    \includegraphics[width=\linewidth]{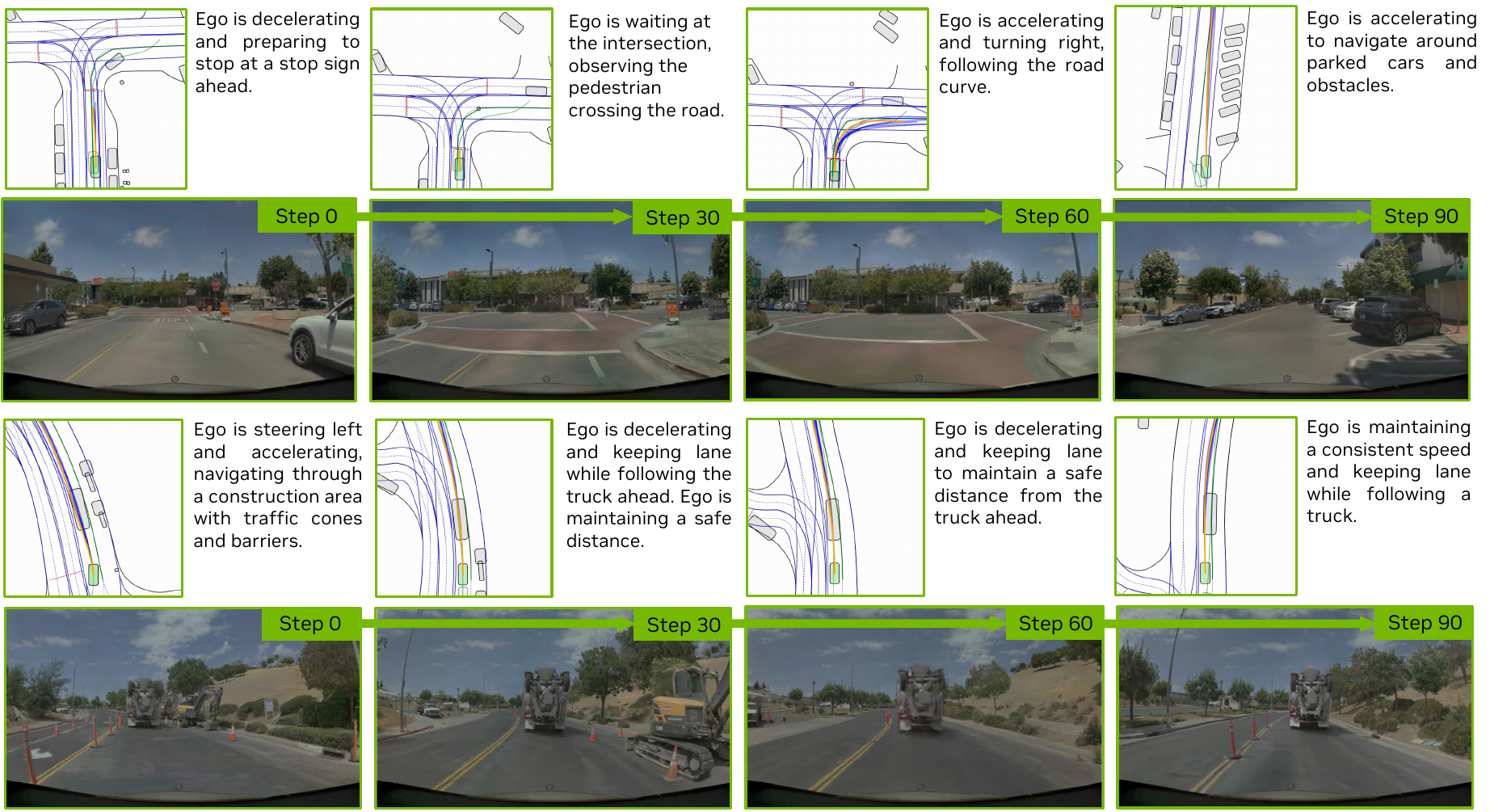}
    \caption{Examples of closed-loop evaluation in AlpaSim~\citep{alpasim2025}. The top row presents an intersection scenario, whereas the bottom row illustrates a construction scenario.}
    \label{fig:cot_closed_loop_results}
\end{figure}

\textbf{Closed-Loop Improvements.}
As shown in \cref{tab:cl_results}, \reasoningvlashort achieves a 35\% reduction in close encounter rate (11\% vs 17\%) compared to the trajectory-only baseline, and a comparable off-road rate (4\% vs 3\%). The overall AlpaSim score improves from 0.38 to 0.50, demonstrating that reasoning-based decision making improves safety in dynamic closed-loop scenarios. \cref{fig:cot_closed_loop_results} presents two qualitative examples that demonstrate that our model can successfully perform closed-loop driving in challenging scenarios within AlpaSim.

\subsection{Improvements of Reasoning, Consistency, and Safety via RL Post-Training}
\label{subsec::alignment_results}

\begin{table}[t]
    \centering
    \caption{Closed-loop evaluation results in AlpaSim~\citep{alpasim2025}. All models are evaluated without route information across 75 challenging scenarios. Baseline refers to the trajectory-only model fine-tuned on CoC training data without reasoning.}
    \label{tab:cl_results}
    \resizebox{\linewidth}{!}{
    \begin{tabular}{lccccc}
        \toprule
                  & \textbf{Close Encounter Rate} 
                  & \textbf{Close Encounter Rate} 
                  & \textbf{Off-Road Rate} 
                  & \textbf{AlpaSim Score} 
                  & \textbf{AlpaSim Score}\\
        \textbf{Model} 
                & \textbf{all  (\%) $\downarrow$} 
                & \textbf{at-fault (\%)  $\downarrow$} 
                & \textbf{(\%) $\downarrow$ } 
                & \textbf{all $\uparrow$}
                & \textbf{at-fault $\uparrow$}\\
        \midrule
        Baseline          & 17.0$\pm$3.0      & 6.0$\pm$1.0   & 3.0$\pm$2.0  &  0.38$\pm$0.04  & 0.86$\pm$0.11 \\
        \rowcolor{mygreen}
        \reasoningvla    & 11.0$\pm$2.0      & 5.0$\pm$3.0   & 4.0$\pm$3.0 & 0.50$\pm$0.08  & 0.87$\pm$0.18 \\
        \bottomrule
    \end{tabular}}
\end{table}

While SFT on \datashortname data enables the model to jointly generate reasoning traces and actions, it does not guarantee that these traces are causally grounded or that the resulting actions faithfully reflect the reasoning or align with human driving norms.
To address this gap, we apply RL-based post-training to simultaneously improve reasoning quality, reasoning-action consistency, and trajectory quality (see \cref{sec:rl_alignment} for methodology details).
In this section, we post-train a 0.5B \reasoningvlashort model fine-tuned on \datashortname data, and demonstrate the impact of different reward components on model behavior.

\begin{figure}[h]
    \centering
    \includegraphics[width=0.95\linewidth]{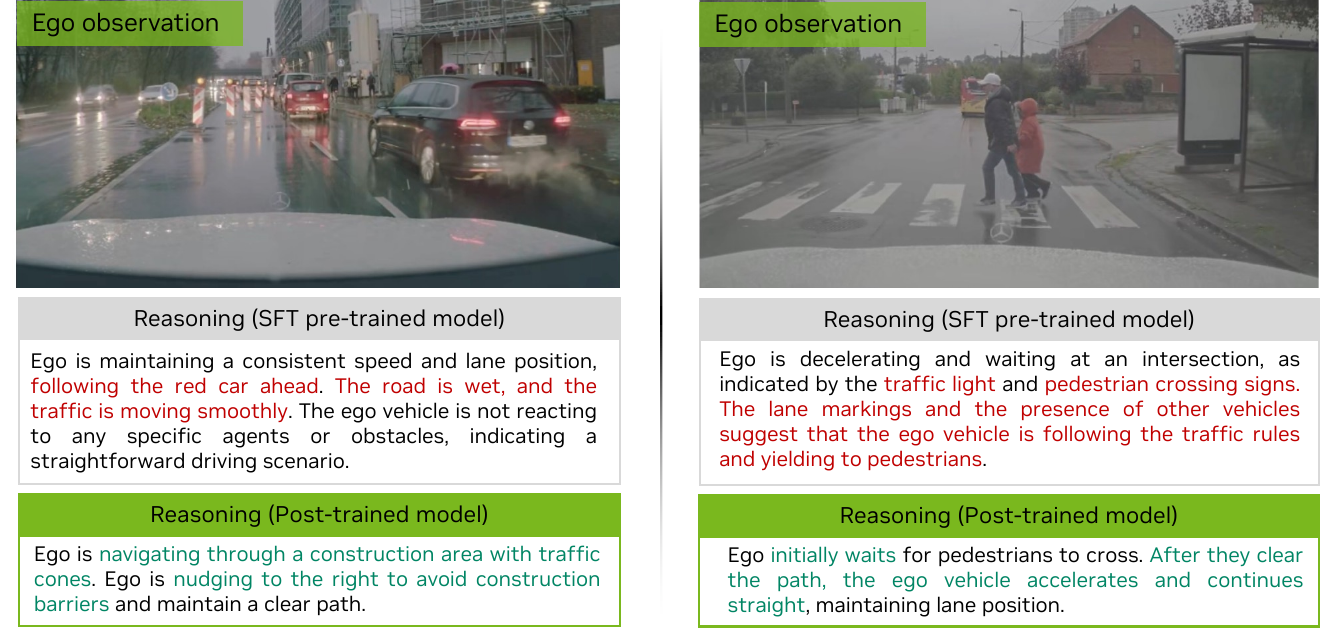}
    \caption{Post-training with the reasoning reward improves causal understanding and contextual reasoning in driving scenarios. \textbf{Left}: The base model overlooks construction barriers and fails to initiate evasive action, while the post-trained model correctly reasons that the ego should nudge right to avoid obstacles. \textbf{Right}: The base model misses that pedestrians are clearing the path, whereas the post-trained model correctly reasons that it is safe for the ego vehicle to accelerate.}
    \label{fig:reasoning_reward_demo}
\end{figure}

\textbf{The Value of Learning from LRM Feedback}. To ensure that the model’s reasoning traces are not only fluent but also causally grounded and contextually accurate, we introduce a reasoning reward derived from LRM feedback (more details are in \cref{sec:rl_alignment}). This reward provides a continuous evaluation signal that measures the logical consistency and causal correctness of each generated reasoning trace with respect to the driving scene. Specifically, the average reasoning score of the most-likely rollout among six generations improves by approximately 45\% (3.1$\rightarrow$4.5) when the reasoning reward is applied. 
In \cref{fig:reasoning_reward_demo}, we illustrate two qualitative examples showcasing the model’s behavioral differences before and after post-training. In the left scenario, the ego vehicle approaches a construction site. The most-likely mode generated by the SFT-pretrained model overlooks the construction barriers and describes the scene as a normal driving situation, failing to recognize the need for evasive behavior. After post-training, however, the model’s reasoning correctly attends to the construction area and explains that the ego vehicle should nudge right to avoid obstacles. Similarly, in the right scenario, two pedestrians are about to clear the path. The most-likely mode generated by the SFT-pretrained model overlooks this contextual cue and fails to anticipate that the ego vehicle should prepare to accelerate. After post-training, the model correctly recognizes that the pedestrians are exiting the drivable area and reasons that it is safe for the ego vehicle to resume motion.

\begin{figure}[t]
    \centering
    \includegraphics[width=0.95\linewidth]{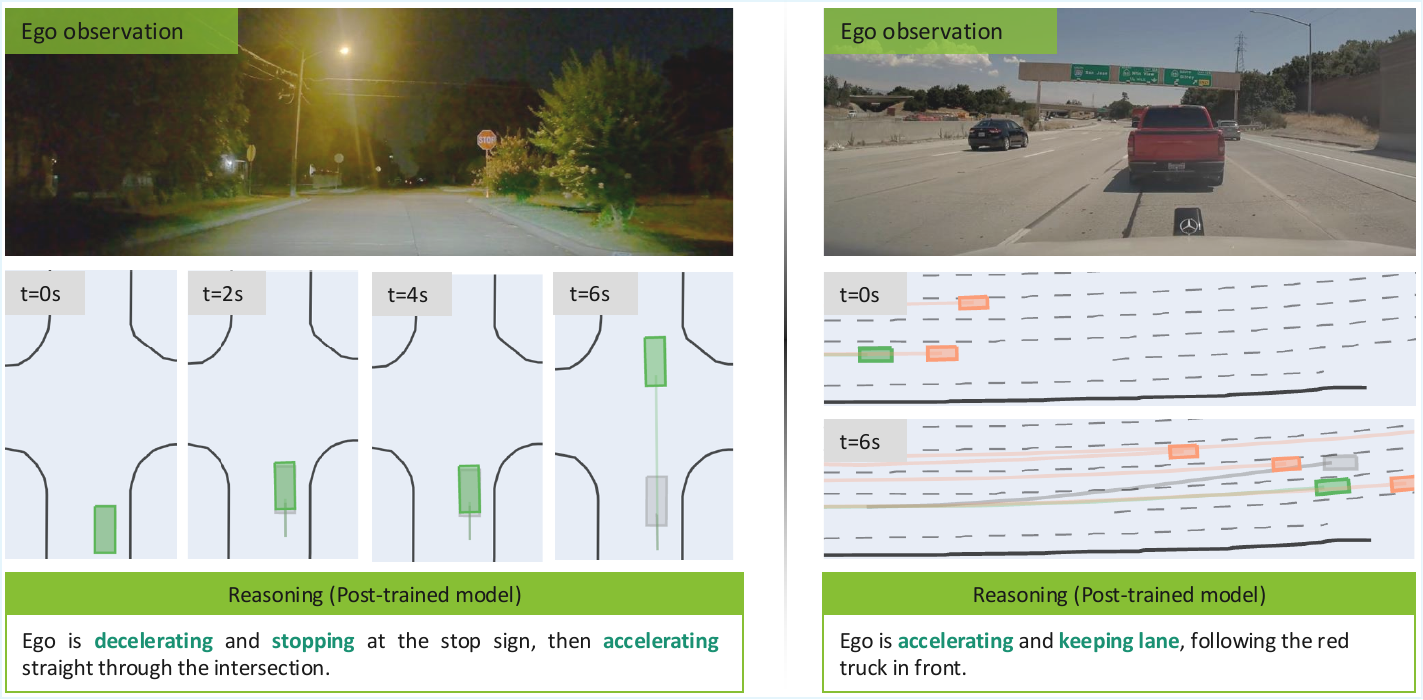}
    \caption{Post-training with the reasoning–action consistency reward improves motion fidelity.
Grey motion denotes the most-likely rollout from the SFT-pretrained base model, and green motion denotes the most-likely rollout from the post-trained model. The orange motions denote the obstacles' motion replay.
\textbf{Left}: The base model (grey) stops halfway and fails to resume motion, even though its reasoning trace correctly instructs the ego vehicle to accelerate after stopping. The post-trained model (green) executes the full causal sequence: decelerating, stopping, and accelerating once the intersection is clear. \textbf{Right}: When the reasoning instructs the ego vehicle to follow a lead vehicle, the post-trained model's generated motion maintains appropriate speed and lane position in accordance with its reasoning trace (``accelerating and keeping lane''), whereas the base model's generated motion changes the lane, drifting from the intended plan.}
\label{fig:reasoning_consistency_reward_demo}
\end{figure}

\begin{table}[t]
    \centering
    \caption{Improvements from RL-based post-training. We evaluate the impact of RL-based post-training on the model’s reasoning, consistency, and motion quality. Metrics are computed from the most-likely rollout among six generated rollouts to assess how RL alignment influences the model's generation distribution. We measure ADE, reasoning quality graded by the large reasoning critic (\cref{sec:rl_alignment_reward}), reasoning–action consistency, and close encounter rate. Evaluations are conducted on the full CoC dataset introduced in \cref{subsec::reasoning_results}. We compare four configurations: the SFT-pretrained base model and three RL post-training variants incorporating different combinations of reasoning, consistency, and safety rewards.
    }
    \label{tab:post-training-result}
    \resizebox{\textwidth}{!}{%
    \begin{tabular}{lcccc}
        \toprule
         \textbf{Training strategy} & \textbf{ADE} $\downarrow$ & \textbf{Reasoning Grading} $\uparrow$ & \textbf{Reasoning–Action Consistency Score} $\uparrow$ & \textbf{Close Encounter Rate ($\%$)} $\downarrow$ \\
        \midrule
        SFT & 2.12m & 3.1 & 0.62 & 6.9\\
        SFT + RL ($r_{\text{reason}}$) & 2.19m & 4.5 &  0.53 & 5.8\\
        SFT + RL ($r_{\text{reason}}$ + $r_{\text{consistency}}$) &  1.92m&  4.5& 0.85  & 6.2\\
        \rowcolor{mygreen} SFT + RL ($r_{\text{reason}}$ + $r_{\text{consistency}}$ + $r_{\text{safety}}$) & 1.94m & 4.4 & 0.83 & 3.7\\
        \bottomrule
    \end{tabular}%
    }
\end{table}

\textbf{The Value of Enforcing Reasoning-Action Consistency}. Interestingly, when the post-training stage optimizes solely for the reasoning reward, the reasoning score indeed improves; however, both the ADE metric and reasoning–action consistency degrade compared to the base model. This indicates that optimizing for reasoning quality alone can lead to ungrounded or overconfident reasoning, where the model produces fluent but causally disconnected explanations that fail to translate into coherent actions. The consistency reward is therefore crucial for anchoring reasoning to physically realizable behaviors, ensuring that improvements in interpretability do not come at the expense of control fidelity. Specifically, when jointly optimizing both the reasoning and consistency rewards, the post-trained model achieves a 9.4\% reduction in most-likely mode ADE (2.12m$\rightarrow$1.92m), a 45\% improvement in the reasoning score (3.1$\rightarrow$4.5), and a 37\% increase in reasoning–action consistency (0.62$\rightarrow$0.85).
These results demonstrate that the two reward components are complementary: the reasoning reward enhances interpretability and causal grounding, while the consistency reward ensures that the generated reasoning translates into faithful and more accurate motion behaviors.
In \cref{fig:reasoning_consistency_reward_demo}, we present two qualitative examples illustrating how post-training improves the model’s motion fidelity. When the model reasons “decelerate, stop, and then accelerate at a stop sign,” the aligned model produces actions that faithfully follow this causal sequence (decelerating smoothly, coming to a complete stop, and accelerating only once the intersection is clear), whereas the SFT-pretrained model tends to stop halfway and never resume motion.

\textbf{The Value of Imposing a Safety Reward}. While reasoning and consistency rewards improve interpretability and causal grounding, they do not explicitly constrain the model to produce safe motion trajectories. To ensure physical safety, we introduce a safety reward that penalizes unsafe or physically implausible trajectories during post-training. Empirically, adding the safety reward further reduces the close encounter  rate and stabilizes trajectory generation without compromising reasoning quality. As shown in~\cref{tab:post-training-result}, the full reward configuration achieves the lowest close encounter  rate while maintaining improvements in ADE and reasoning–action consistency.

\subsection{Public Benchmark Evaluation}
\label{subsec::public_benchmark}

To enable reproducible evaluation and community comparison, we evaluate the publicly released \reasoningvla model on the PhysicalAI-AV dataset~\citep{nvidia2025avdata} and the AlpaSim public scenario set~\citep{alpasim2025}. \reasoningvla-10B leverages Cosmos-Reason~\citep{nvidia2025cosmosreason1physicalcommonsense} as the VLM backbone with a 2B parameter diffusion-based trajectory decoder, while \reasoningvla-0.5B uses a smaller backbone for comparison. Both models generate Chain-of-Causation reasoning traces alongside trajectory predictions.

\begin{table}[t]
    \centering
    \caption{Evaluation of \reasoningvla on the public PhysicalAI-AV dataset~\citep{nvidia2025avdata}. Open-loop results are evaluated on 644 examples from the PhysicalAI-AV evaluation set. Closed-loop results are evaluated on 920 scenarios from the PhysicalAI-AV NuRec dataset~\citep{nvidia2025nurecavdata} using AlpaSim~\citep{alpasim2025}. All models predict CoC reasoning traces and trajectories. Closed-loop metrics are at-fault.}
    \label{tab:public_benchmark_results}
    \resizebox{\linewidth}{!}{
    \begin{tabular}{l|c|ccc}
        \toprule
        \multirow{2}{*}{\textbf{Model}} & \textbf{Open-Loop} & \multicolumn{3}{c}{\textbf{Closed-Loop (AlpaSim)}} \\
        \cmidrule(lr){2-2} \cmidrule(lr){3-5}
        & \textbf{minADE$_6$@6.4s}$\downarrow$ & \textbf{Close Encounter Rate $\downarrow$ (\%)} &  \textbf{Off-Road Rate $\downarrow$ (\%)} & \textbf{AlpaSim Score}$\uparrow$ \\
        \midrule
        \reasoningvla-0.5B & 0.913 & 9.0$\pm$1.0 & 19.0$\pm$0.0 & 0.35$\pm$0.01 \\
        \rowcolor{mygreen}
        \reasoningvla-10B  & 0.849 & 4.0$\pm$0.0 & 16.0$\pm$1.0 & 0.72$\pm$0.02 \\
        \bottomrule
    \end{tabular}}
\end{table}

\textbf{Open-Loop Results.} We evaluate both models on 644 held-out examples from the PhysicalAI-AV dataset. As shown in \cref{tab:public_benchmark_results}, \reasoningvla-10B achieves a minADE$_6$ of 0.849m at 6.4s, a 7.0\% improvement over \reasoningvla-0.5B (0.913m), demonstrating the benefits of scaling the VLM backbone.

\textbf{Closed-Loop Results.} We further evaluate both models on the AlpaSim public scenario set comprising 920 challenging driving scenarios. As shown in \cref{tab:public_benchmark_results}, \reasoningvla-10B achieves substantial improvements across all closed-loop metrics: a 16\% reduction in off-road rate (16\% vs 19\%), 
a 55\% reduction in close encounter rate (4\% vs 9\%), 
and more than 2$\times$ improvement in AlpaSim score (0.72 vs 0.35). These results demonstrate that scaling model capacity significantly enhances the model's ability to handle complex driving scenarios in closed-loop simulation.

\begin{figure}[t]
    \centering
    \includegraphics[width=0.7\linewidth]{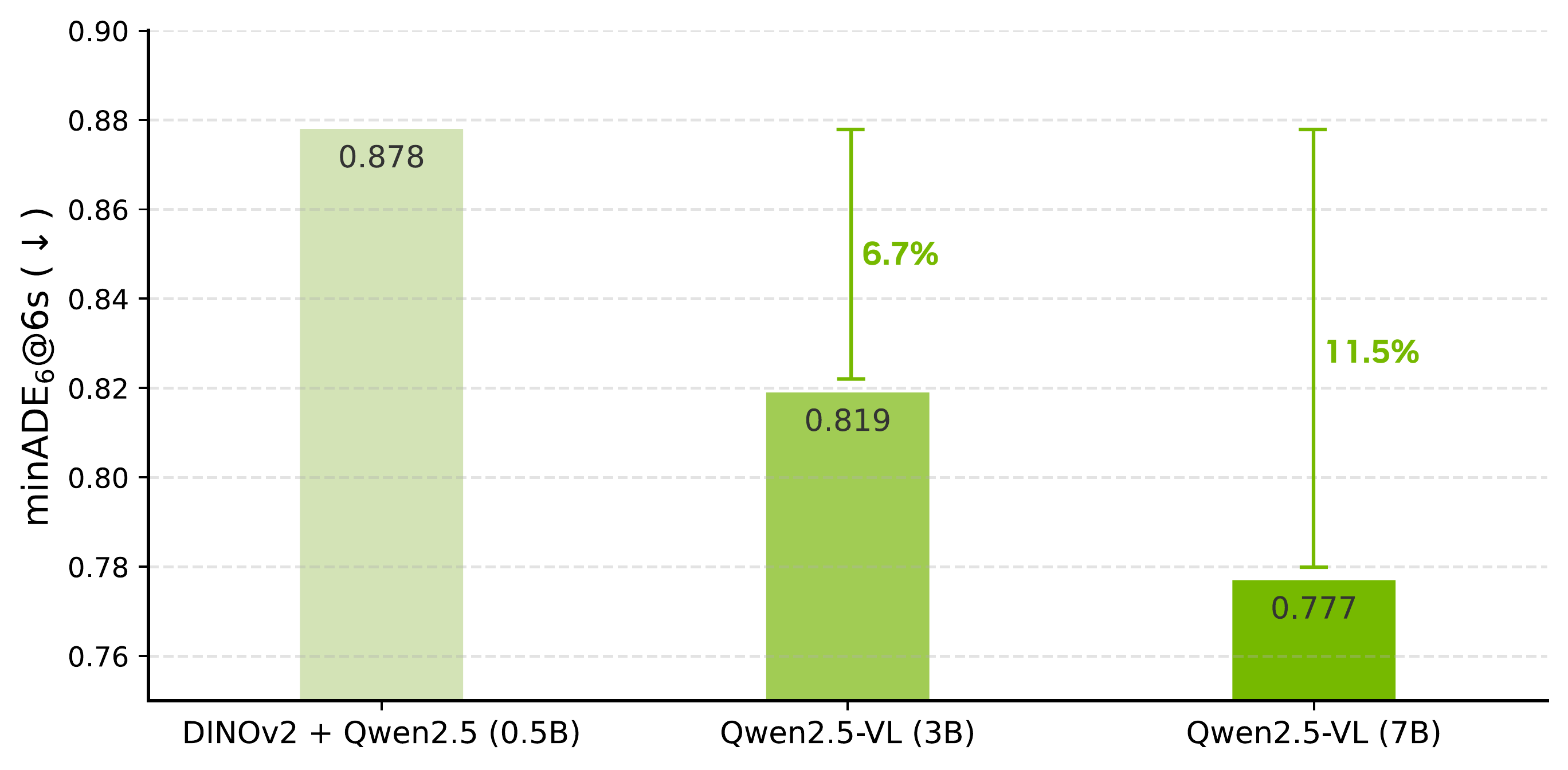}
    \vspace{-0.2cm}
    \caption{Impact of VLM backbone size on open-loop driving performance. All models are evaluated on $\mathcal{D}_{\text{overall}}$ with the same training data and hyperparameters.}
    \label{fig:vlm_scaling_results}
\end{figure}

\subsection{Ablation: VLM Backbone Selection}
\label{subsec::abl_backbone}

The choice of VLM backbone is critical for \reasoningvla's performance. In this section, we investigate two complementary aspects: the impact of model scale and the benefits of Physical-AI-focused pre-training. Together, these ablations demonstrate that both model capacity and domain-relevant pre-training are essential for strong driving performance.

\subsubsection{Model Size Ablation}
To investigate the impact of model capacity on driving performance, we first conduct baseline scaling experiments using general-purpose VLMs. Specifically, we evaluate three variants of our architecture with different backbone sizes: 0.5B, 3B, and 7B parameters. The 0.5B model uses a DINOv2~\citep{oquab2023dinov2} vision encoder combined with the Qwen2.5-0.5B~\citep{qwen2.5} language model, while the 3B and 7B models leverage Qwen2.5-VL-3B~\citep{bai2025qwen2} and Qwen2.5-VL-7B~\citep{bai2025qwen2}, respectively. For this ablation study, all variants are trained on identical data with a reduced training budget compared to our main models, and evaluated on $\mathcal{D}_{\text{overall}}$ held-out test set without route information, using the minADE$_6$ metric over a 6.4s horizon.

As shown in \cref{fig:vlm_scaling_results}, we observe consistent improvements in open-loop performance as model size increases. The 7B model achieves a reduction of 11\% in minADE$_6$ compared to the baseline of 0.5B, demonstrating that scaling the vision-language backbone enables better scene understanding and trajectory prediction. While these results confirm the importance of model capacity, they are based on general-purpose VLMs without domain-specific pre-training. As we demonstrate in ~\cref{sec::exp::cosmos_reason}, incorporating Physical AI-focused pre-training (via Cosmos-Reason, \cref{sec::exp::cosmos_reason}) yields substantial further improvements, which is why our final \reasoningvla models adopt Cosmos-Reason as their backbone.

\subsubsection{Data Scaling}
\label{subsec::data_scaling}
Complementary to model scaling, we investigate how training data scale affects driving performance when model architecture and training budget are held constant. We train the 0.5B model on varying amounts of data: 100k, 200k, 500k, 1M, and 2M video segments, keeping the total number of training steps fixed across all experiments.
\begin{figure}[t]
    \centering
    \includegraphics[width=0.7\linewidth]{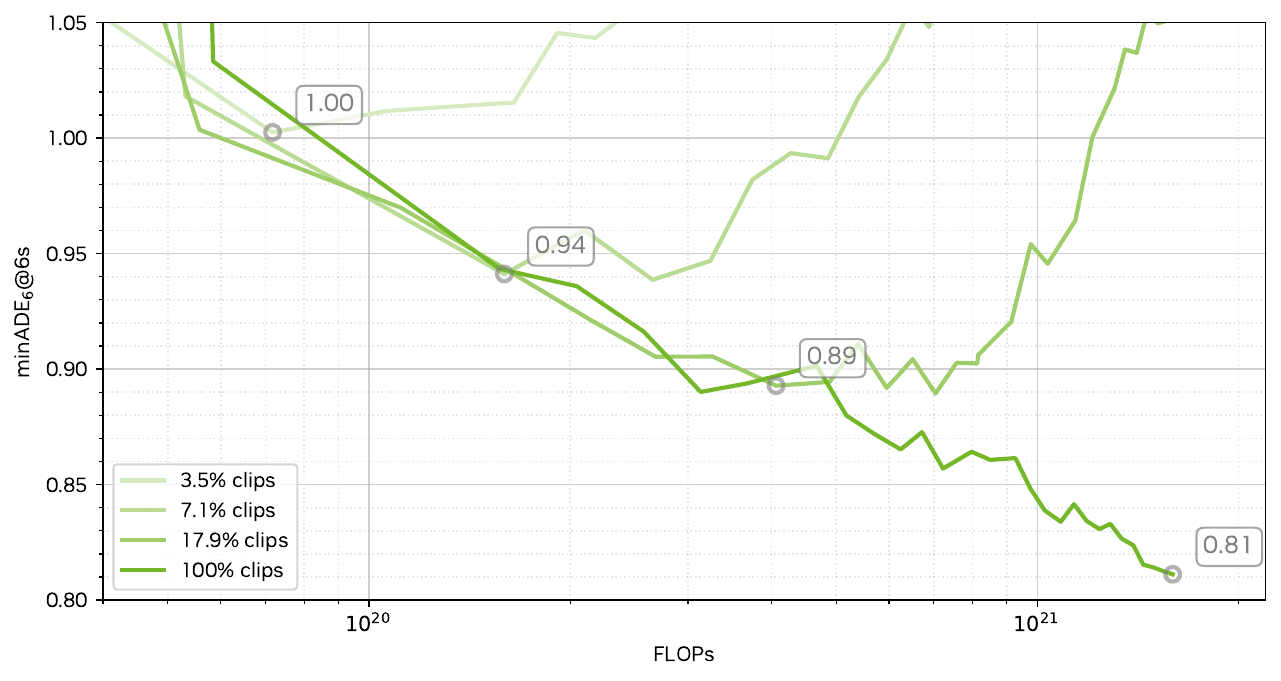}
    \vspace{-0.2cm}
    \caption{Impact of training data scale on open-loop driving performance. All models use the 0.5B architecture with identical hyperparameters and fixed total training steps. Models are evaluated on $\mathcal{D}_{\text{overall}}$ held-out test set. 
    }
    \label{fig:data_scaling_results}
\end{figure}

As shown in \cref{fig:data_scaling_results}, performance consistently improves with increased data scale, demonstrating the value of data diversity for autonomous driving. The 100k model exhibits clear overfitting (1.111m without early stopping; 1.016m with early stopping). Scaling to 500k achieves 0.880m (13.4\% improvement over 100k), while 2M achieves the best performance at 0.874m (14.0\% improvement). These results, together with the model size ablation in the previous subsection, demonstrate that both model capacity and data scale are effective dimensions for improving driving performance, underscoring their complementary roles in achieving robust autonomous driving systems.

\subsubsection{Cosmos-Reason Physical AI Capabilities}
\label{sec::exp::cosmos_reason}
While the scaling experiments above demonstrate the importance of model capacity, they do not address a critical question: given a fixed model size, does domain-specific pre-training matter? As described in \cref{sec::model}, \reasoningvla adopts Cosmos-Reason~\citep{nvidia2025cosmosreason1physicalcommonsense} as its VLM backbone, specifically post-trained on Physical AI data including driving scenarios. To validate this architectural choice and demonstrate that Physical-AI-focused pre-training enhances driving-specific understanding beyond what scale alone provides, we evaluate Cosmos-Reason against comparable 7B-scale general-purpose VLMs on public driving benchmarks.

\textbf{LingoQA Benchmark.} \cref{tab:lingoqa} presents zero-shot evaluation results on the LingoQA benchmark~\citep{marcu2024lingoqa}, which assesses vision-language models on driving scene understanding. Our Cosmos-Reason-7B model achieves 66.2\% accuracy, outperforming various VLMs including GPT-4V (59.6\%), Qwen2-VL-7B (52.6\%), Qwen2.5-VL-7B (62.2\%), InternVL3.5-8B (58.6\%), and DeepSeek-VL-7B (46.4\%). This improvement over the baselines demonstrates that Physical-AI-focused SFT significantly improves scene understanding capabilities for autonomous driving contexts, complementing the benefits of model scaling shown in \cref{fig:vlm_scaling_results}.

\begin{table}[tb!]
\centering
\caption{Zero-shot accuracy of various VLMs on the LingoQA benchmark~\citep{marcu2024lingoqa}. Our Cosmos-Reason-7B model outperforms all baselines.}
\resizebox{0.95\linewidth}{!}{%
\begin{tabular}{l|cccccc}
\toprule
Model & GPT-4V & Qwen2-VL-7B & Qwen2.5-VL-7B & InternVL3.5-8B & DeepSeek-VL-7B & Ours \\ \midrule
Lingo-Judge & 59.6 & 52.6 & 62.2 & 58.6 & 46.4 & \textbf{66.2} \\ \bottomrule
\end{tabular}}
\label{tab:lingoqa}
\end{table}

These results confirm that \textit{both} model capacity \textit{and} domain-specific pre-training are essential for strong driving performance. This motivates our choice of Cosmos-Reason as the backbone for \reasoningvla, providing a strong foundation with Physical AI capabilities that general-purpose VLMs may otherwise not have.

\subsection{Ablation: Action Modality Injection}
\label{subsec::abl_action}

We demonstrate the effectiveness of adopting a continuous action representation governed by unicycle dynamics with flow matching in \cref{tab:traj_decoder_results}. 
Specifically, we compare a baseline model trained to auto-regressively predict 6 discrete trajectory tokens against a model of identical size and training data that decodes trajectories via flow matching. 
The discrete trajectory tokenizer in the baseline auto-regressive model is pre-trained via VQGAN~\citep{esser2021taming}, which minimizes the number of output discrete tokens to reduce the auto-regressive decoding latency while maintaining low reconstruction error.
During inference, we set $\delta_t = 0.2$, i.e., 5 steps, in flow matching to reduce latency with negligible performance degradation.
As shown in \cref{tab:traj_decoder_results}, leveraging a dynamically governed continuous action space through flow-matching yields substantial improvements in both open-loop and closed-loop metrics, enhancing comfort and achieving faster inference speed.

\begin{table}[t]
    \centering
    \caption{Comparison on trajectory decoding strategies. The models are trained and evaluated with route signals. 
    The evaluation is on $\mathcal{D}_{\text{overall}}$ to show overall gains. 
    Comfort (Accel) metric measures the percentage of predicted trajectories that are within a comfort range.}
    \label{tab:traj_decoder_results}
    \resizebox{\linewidth}{!}{%
    \begin{tabular}{lcccc}
        \toprule
        \textbf{Strategy} & \textbf{$\minade$@6.4s} $\downarrow$  & \textbf{AlpaSim Score (at fault)} $\uparrow$  & \textbf{Comfort (Accel)} $\uparrow$ & \textbf{Rel. Decode Speed}$\uparrow$  \\
        \midrule
        Auto-Regressive & 0.6811 & 0.59 $\pm$ 0.17 & 44.05\% & 1.00$\times$\\ %
        \rowcolor{mygreen}
        Flow Matching & 0.6440 & 1.27 $\pm$ 0.34 & 97.38\% & 1.16$\times$\\ %
        \bottomrule
    \end{tabular}
    }
\end{table}

\subsection{Ablation: Efficient Vision Encoding}
\label{sec::exp::visenc}
As discussed in~\cref{sec::vision_encoding}, there are alternative methods for vision encoding that can be more efficient than the default single-image tokenizer in terms of tokens needed to represent multi-camera video inputs. To compare approaches, we choose a 4-camera setup, vary the vision encoder, and compare the resulting end-to-end model's open-loop driving quality via minADE$_6$ relative to the baseline.

As can be seen in~\cref{tab:visenc_results}, the triplane-based multi-camera tokenizer from~\citet{ivanovic2025efficient} achieves nearly identical minADE$_6$ values as the baseline, while only adding 6.3M parameters and reducing sensor token counts by $3.6\times$. Flex~\citep{yang2025flex} is able to achieve more drastic improvements, with a token compression of up to $20\times$ while only adding 61.6M parameters to the overall driving model and matching the driving quality of the baseline.

\reasoningvlashort{} adopts single-image tokenization by default, as the optimal strategy can vary with the number of cameras, temporal frames, and camera resolutions. For example, a small number of cameras and short histories will favor single-image tokenization, more cameras and short histories will favor triplanes~\citep{ivanovic2025efficient}, and more cameras and long history sequences will favor Flex~\citep{yang2025flex}.

\begin{table}[t]
    \centering
    \caption{Relative comparison of different efficient vision encoding strategies on $\mathcal{D}_{\text{overall}}$.}
    \label{tab:visenc_results}
    \begin{tabular}{lcrc}
        \toprule
        \textbf{Vision Encoder} & 
        \textbf{Added Parameters $\downarrow$} & 
        \textbf{Tokens per Image $\downarrow$} & \textbf{Rel. minADE$_6$} $\downarrow$  \\
        \midrule
        Baseline & 0 & $160$ $(1.0\times)$   & $0\%$ \\
        \midrule
        \multirow{2}{*}{\shortstack[l]{Triplane \\ \citep{ivanovic2025efficient}}} &     \multirow{2}{*}{6.3M}     & $104$ $(1.5\times)$    & $-3\%$ \\
         &          & $45$ $(3.6\times)$     & $+4\%$ \\
        \midrule
        \multirow{4}{*}{\shortstack[l]{Flex \\ \citep{yang2025flex}}} & \multirow{4}{*}{61.6M} & $50$ $(3.2\times)$ & $-3\%$ \\
         &  & $32$ $(5.0\times)$ & $-3\%$ \\
         &  & $16$ $(10\times)$ & $-2\%$ \\
         &  & $8$ $(20\times)$ & $-2\%$ \\
        \bottomrule
    \end{tabular}
\end{table}

\subsection{On-Vehicle Road Tests}
\begin{figure}[t]
    \centering
    \includegraphics[width=\linewidth]{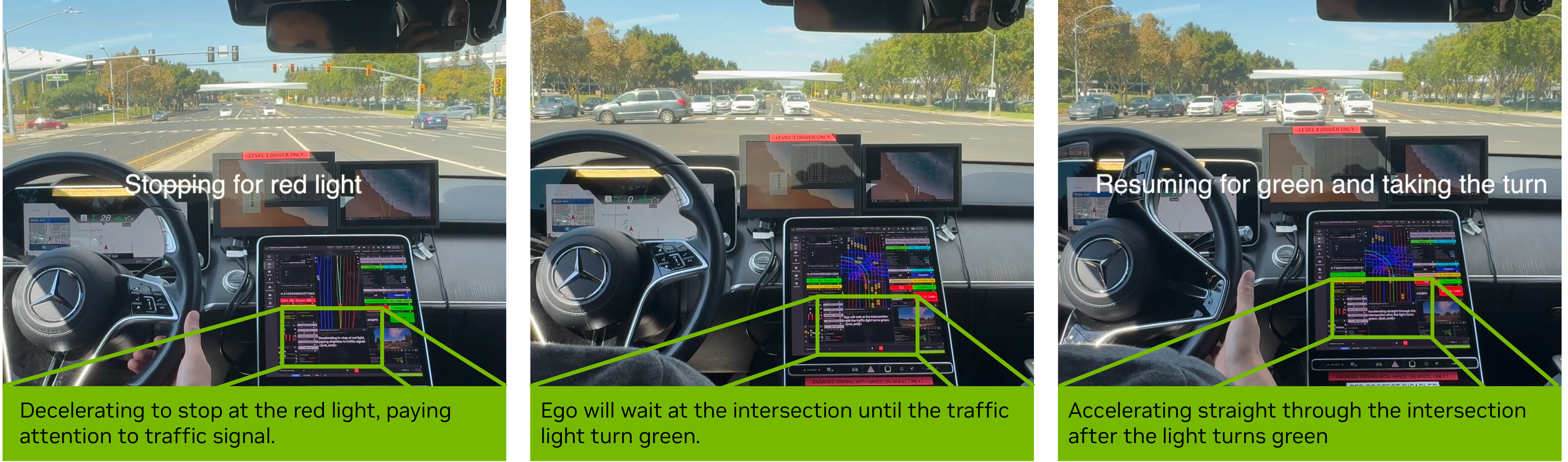}
    \caption{On-vehicle road test showing that \reasoningvlashort generates a reasoning trace in an intersection scenario. The ego vehicle first decelerates to stop due to the red light, then waits for the traffic signal and finally resumes when the light turns green and takes the turn.}
    \label{fig:road_test}
\end{figure}

To validate the real-world deployment capability of \reasoningvlashort, we deployed the model in a test vehicle and conducted road testing in urban driving environments. The vehicle successfully navigated complex urban scenarios without human intervention, demonstrating the model's ability to handle real-world driving conditions beyond simulation.
\cref{fig:road_test} shows an intersection where \reasoningvlashort accurately identifies the traffic situation and produces clear and concise reasoning traces that lead to appropriate driving actions. These tests confirm that simulation improvements are transferred successfully to real-world autonomous driving scenarios.

\textbf{Real-Time Inference Performance.} A critical requirement for on-vehicle deployment is real-time inference capability. We benchmark \reasoningvlashort on an NVIDIA RTX 6000 Pro Blackwell platform, achieving an end-to-end inference latency of 99ms, within the real-time requirements for autonomous driving (typically 100ms). \cref{tab:inference_time} provides a detailed breakdown of the inference pipeline, comparing our approach against alternative design choices. The prefilling stage processes the visual tokens and route information through the transformer layers to generate the key-value cache, which is then used during both reasoning and trajectory decoding.

\begin{table}[t]
    \centering
    \caption{Inference runtime breakdown on an NVIDIA RTX 6000 Pro Blackwell. \reasoningvla achieves real-time performance (99ms) by combining flow-matching-based trajectory decoding with efficient vision encoding.}
    \label{tab:inference_time}
    \resizebox{\textwidth}{!}{%
    \begin{tabular}{lccccc}
        \toprule
        \textbf{Model Configuration} & \textbf{Vision Encoder} & \textbf{Prefilling} & \textbf{Reasoning Decoding} & \textbf{Trajectory Decoding} & \textbf{Total} \\
        \midrule
        Baseline (trajectory-only, flow matching) & 3.43ms & 16.54ms & -- & 8.75ms (5 steps) & 29ms \\
        \rowcolor{mygreen}
        \reasoningvla (ours, flow matching) & 3.43ms & 16.54ms & 70ms (40 tokens) & 8.75ms (5 steps) & \textbf{99ms} \\
        \reasoningvla (auto-regressive traj) & 3.43ms & 16.54ms & 70ms (40 tokens) & 222ms (127 tokens) & 312ms \\
        \bottomrule
    \end{tabular}%
    }
\end{table}

\section{Conclusion}
\label{sec::conclusion}

In this work, we present \textbf{\reasoningvla (\reasoningvlashort)}, a vision-language-action model that integrates structured chain-of-thought reasoning capabilities with trajectory prediction to enhance autonomous driving performance, particularly in long-tail, safety-critical scenarios. To enable the model to generate causally-grounded reasoning, we introduce the \textbf{\datafullname (\datashortnamenosp)} dataset, constructed through a hybrid labeling pipeline that combines large-scale auto-labeling with humans in the loop. We further align reasoning with action through RL, ensuring that the generated reasoning traces are consistent with the executed driving behaviors. Our comprehensive evaluations across open-loop metrics, closed-loop simulation, and ablation studies demonstrate that \reasoningvlashort achieves consistent improvements over end-to-end baselines, with particularly pronounced gains on challenging scenarios involving complex agent interactions.

\mypara{Future Work.} Several promising research directions remain open. First, \textit{policy structuring}: while our flow-matching-based trajectory decoder provides kinematically feasible outputs, exploring hierarchical policy architectures that decompose high-level meta-actions into structured motion primitives could further improve interpretability and efficiency. Second, \textit{reasoning on demand}: our current architecture generates reasoning traces for every input; future work could investigate adaptive mechanisms that selectively invoke reasoning only for safety-critical or ambiguous scenarios, enabling more efficient inference-time computation allocation similar to recent advances in test-time scaling~\citep{yao2023tree,openaio1}; Third, \textit{auxiliary task integration}: while \reasoningvlashort focuses on trajectory prediction and causal reasoning, incorporating complementary self-supervised objectives, such as depth estimation, scene flow prediction, or 3D Gaussian Splatting representations, could improve the visual backbone's semantic understanding; Fourth, \textit{world model integration}: our current approach predicts actions from observed states; incorporating learned world models could enable forward simulation and counterfactual reasoning, improving robustness in dynamic scenarios.

\textbf{Open Source Release.} We release \reasoningvla-10B model weights at \url{https://huggingface.co/nvidia/Alpamayo-R1-10B} and inference code at \url{https://github.com/NVlabs/alpamayo}. The model is evaluated on the PhysicalAI-AV dataset~\citep{nvidia2025avdata} and the AlpaSim public scenario set~\citep{alpasim2025}, enabling reproducible benchmarking by the research community.

\clearpage
\appendix

\clearpage
\section{Contributors and Acknowledgments}
\label{sec::contributors}

\subsection{Core Contributors}

\noindent
Yulong Cao,
Tong Che,
Yuxiao Chen,
Wenhao Ding,
Boris Ivanovic,
Peter Karkus,
Boyi Li,
Tsung-Yi Lin,
Patrick Langechuan Liu,
Zhijian Liu,
Jason Lu,
Wenjie Luo,
Marco Pavone,
Ran Tian,
Yan Wang,
Xinshuo Weng,
Tianjun Xiao,
Xiaodong Yang,
Yurong You,
Xiaohui Zeng.

\vspace{2mm}
\textbf{Data \& Benchmarks:} TX, XW, YC, WD, YW curated autonomous driving datasets and benchmarks.\\
\textbf{Labeling Pipeline:} XW, YC, WD, BL, XY, YW developed the reasoning trace labeling pipeline and the infrastructure. \\
\textbf{Training Infrastructure:} YY, WL, YW, WD built the supervised fine-tuning infrastructure; TC, RT, WL built the reinforcement learning infrastructure. \\
\textbf{Vision Encoding:} BI, YW developed the vision encoder. \\
\textbf{Action Decoding:} YY, YC built the flow-matching trajectory decoder. \\
\textbf{Model Training:} YY, WL, YW, WD, JL, ZL, PLL trained the VLA models with supervised fine-tuning; YW, WL, YY, XY, TL, XZ trained the Cosmos-Reason VLM backbone; RT, TC, YW, WL, YY, WD designed the post-training strategy and post-trained models with reinforcement learning; WD, YC designed the data mixture strategy.\\
\textbf{Project Leads:} YW, WL drove the project from concept to completion.\\
\textbf{Program Architect and Project Manager:} MP conceived, coordinated, and guided the overall effort. BI supported MP in coordination and guidance.

\subsection{Contributors}

\noindent
Junjie Bai, 
Ke Chen,
Jenna Diamond,
Yifan Ding,
Liang Feng,
Greg Heinrich,
Jack Huang,
Pinyi Li, 
Dongran Liu,
Ming-Yu Liu,
Leo Yunxiang Mao,
Pavlo Molchanov,
Lindsey Pavao,
Zhenghao Peng,
Mike Ranzinger,
Ed Schmerling,
Shida Shen,
Yunfei Shi,
Sarah Tariq,
Tilman Wekel,
Eric Yang,
Wenyuan Zhang.

\vspace{2mm}
\noindent \textbf{Contributions.} 
ST led the end-to-end development on the production side and provided key input on the data pipeline and model architecture. LP, JD led the human annotation effort. PM, GH, MR trained the vision encoder. 
ML provided Cosmos-Reason model support.
YD processed cosmos AV data into training format.
ZP improved the large-scale SFT training workflow. 
FL, JB provided support for the large-scale RL training infrastructure.
ES curated and preprocessed driving data. 
KC, WZ, JH improved the \datashortname auto-labeling pipeline.
SS developed the LLM-based evaluator for \datashortname reasoning traces.
YS, EY, TW built the \datashortname labeling tools for human labeling. DL, PL and LM were instrumental in conducting the on-vehicle tests and model profiling.

\subsection{Acknowledgments}

We thank Xinzhou Wu and Ali Kani for leadership and strategic support; Sachin Patil for general support in AV model training and deployment; Zhiding Yu, Guilin Liu, Max Li, Song Han, Hongxu Yin, Sifei Liu, and Yu-Wei Chao for valuable discussions on vision-language model training; Jesse Hong for running the \datashortname labeling pipeline; Richard Lin, Zi Wang, Walter Yu for improvements to the \datashortname  auto-labeling pipeline; Anton Mitrokhin, Jacob Kern for improvements to the \datashortname human labeling pipeline;  Martin Peng, Steve Hu, Andy Martin for dataset management and releases; Di Chen, Hanson Xu for help with model deployment; Chao Fang, Shuaijun Chen, and Niral Pathak for on-vehicle deployment support; Charles Vorbach, Zhenyi Zhang, Rachit Shah, Ritaank Tiwari for help with onboard vehicle deployment; Parixit Aghera, Ratin Kumar, Parag Mehendale, Niranjan Avadhanam, Rajath Shetty, Ronan LeToquin, Suraj Das and Ashley Hu for vehicle testing; Sachit Kadle, Annie Feng, and Zheng Lian for closed-loop simulation support; Maximilian Igl, Michael Watson, and Apoorva Sharma for closed-loop experimentation and metric implementations.

\clearpage
\setcitestyle{numbers}
\bibliographystyle{plainnat}
\bibliography{main}

\end{document}